%% file: main.tex
\pdfoutput=1

\documentclass[11pt]{article}

\usepackage[final]{main}

\usepackage{times}
\usepackage{latexsym}

\usepackage[T1]{fontenc}

\usepackage[utf8]{inputenc}

\usepackage{microtype}

\usepackage{inconsolata}

\usepackage{graphicx}

\usepackage{algorithmic}
\usepackage{graphicx}
\usepackage{listings}
\usepackage{textcomp}
\usepackage{xcolor}
\usepackage{hyperref}
\usepackage{xspace}
\usepackage{mdframed}
\usepackage{subcaption}
\usepackage{parcolumns}
\usepackage{wrapfig}
\usepackage{fancyvrb}
\usepackage{fvextra}
\usepackage{multirow}

\lstdefinestyle{highlightlines}{
    moredelim=**[is][\color{red}]{@}{@},
    moredelim=**[is][\color{blue}]{^}{^},
}

\title{AutoParLLM: GNN-guided Context Generation for Zero-Shot Code Parallelization using LLMs}

\author{
 \textbf{Quazi Ishtiaque Mahmud\textsuperscript{1}},
 \textbf{Ali TehraniJamsaz\textsuperscript{1}},
 \textbf{Hung Phan\textsuperscript{1}},
 \textbf{Le Chen\textsuperscript{1}},
\\
 \textbf{Mihai Capotă\textsuperscript{2}},
 \textbf{Theodore Willke\textsuperscript{3}},
 \textbf{Nesreen K. Ahmed\textsuperscript{4}},
 \textbf{Ali Jannesari \textsuperscript{1}}
\\
 \textsuperscript{1}Iowa State University
 \textsuperscript{2}Intel Labs
 \textsuperscript{3}DataStax
 \textsuperscript{4}Cisco AI Research
\\
\textsuperscript{1}{\{\texttt{mahmud,tehrani,hungphd,lechen,jannesar}\}\texttt{@iastate.edu}} \\
\textsuperscript{2}{\texttt{mihai.capota@intel.com}}
\textsuperscript{3}{\texttt{ted.willke@datastax.com}}
\textsuperscript{4}{\texttt{nesahmed@cisco.com}}
}

\begin{document}

\newcommand{\ourtool}{\textsc{AutoParLLM}}
\newcommand{\ourscore}{\textsc{OMPScore}}
\newcommand{\perfograph}{\textsc{PerfoGraph}}
\newcommand{\programl}{\textsc{PrograML} \xspace{}}

\maketitle
\begin{abstract}

In-Context Learning (ICL) has been shown to be a powerful technique to augment the capabilities of LLMs for a diverse range of tasks. This work proposes \ourtool, a novel way to generate context using guidance from graph neural networks (GNNs) to generate efficient parallel codes. We evaluate \ourtool \xspace{} on $12$ applications from two well-known benchmark suites of parallel codes: NAS Parallel Benchmark and Rodinia Benchmark. Our results show that \ourtool \xspace{} improves the state-of-the-art LLMs (e.g., GPT-4) by 19.9\% in NAS and 6.48\% in Rodinia benchmark in terms of CodeBERTScore for the task of parallel code generation. Moreover, \ourtool \xspace{} improves the ability of the most powerful LLM to date, GPT-4, by achieving $\approx$17\% (on NAS benchmark) and $\approx$16\% (on Rodinia benchmark) better speedup. In addition, we propose \ourscore \xspace{} for evaluating the quality of the parallel code and show its effectiveness in evaluating parallel codes. \ourtool \xspace is available at \href{https://github.com/quazirafi/AutoParLLM.git}{https://github.com/quazirafi/AutoParLLM.git}.


\end{abstract}

\input{sections/intro}

\input{sections/motivation}

\input{sections/approach}

\input{sections/exp_results}

\input{sections/use_case}
\input{sections/related_works}

\input{sections/conclusion}

\section*{Limitations}

This work is currently focused on OpenMP-based parallel code generation however we demonstrated that \ourtool \xspace can be easily extended to support other parallel programming models (e.g., OpenACC) as well. However, this study does not consider parallelism opportunities that can be obtained by rewriting the code. 

\section*{Acknowledgement}

We would like to thank NSF for their generous support in funding this project (\#2211982) and (\#2422127). In addition,
we extend our gratitude to Intel Labs and Cisco AI Research for supporting this project. We thank the Research IT team of Iowa State University for providing access to HPC clusters for conducting the experiments of this research project.  


\bibliography{main}

\newpage

\appendix

\section{Appendix}

\subsection{Background Examples} 
\label{sec:appendix-background}

\begin{figure}[h]
    \centering
    \includegraphics[width=1\linewidth]{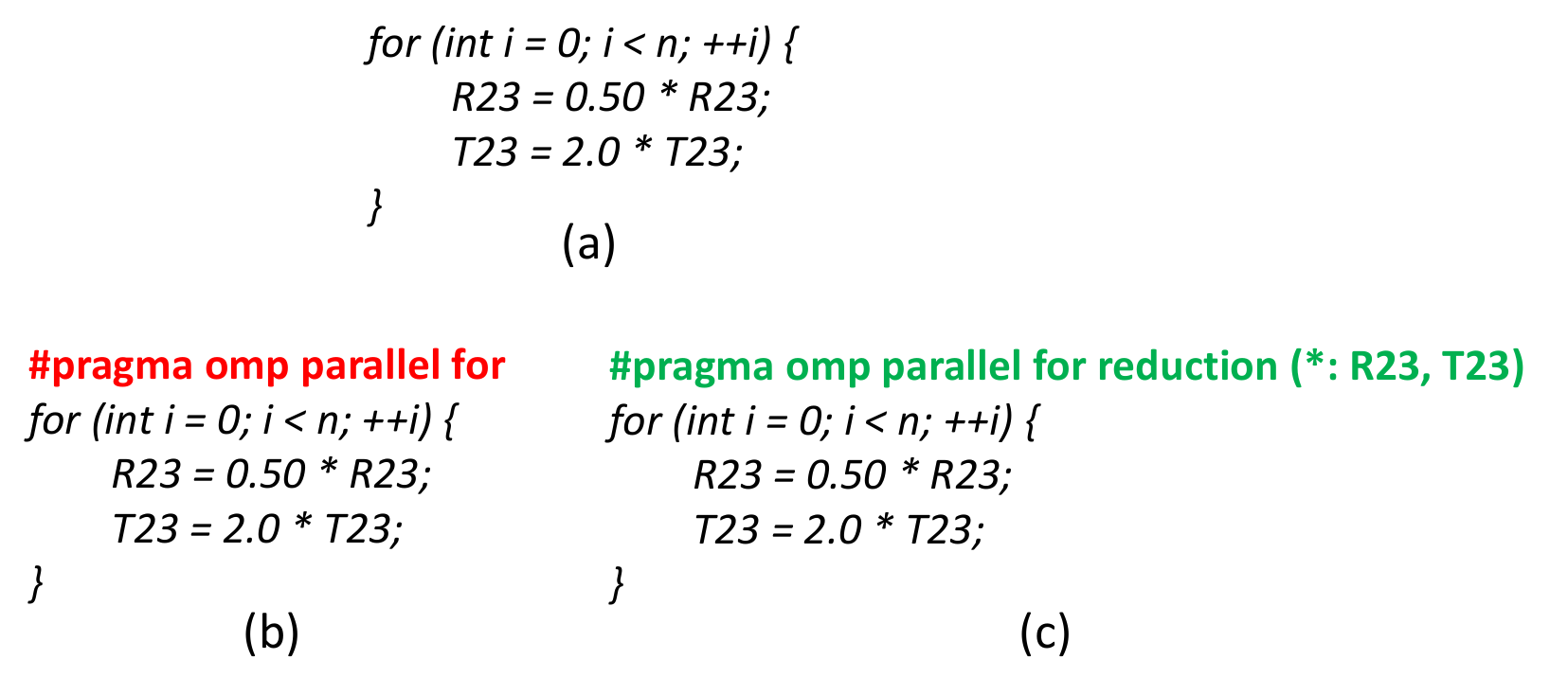}
    \vspace{-25pt}
    \caption{Result of GPT-3.5 and GPT-4 (b) before and, (c) after applying \ourtool \xspace guidance on input code (a). This reduction loop is taken from IS application of NAS Parallel Benchmark}
    \vspace{-10pt}
    \label{fig:abl-study-1}
\end{figure}

\lstset{basicstyle=\footnotesize\ttfamily}

\begin{scriptsize}
\begin{lstlisting}[language=c, label={reduction-ex}, caption=Reduction loop example]
#pragma omp parallel for reduction(+:sum)
for (int i = 0; i < n; ++i) {
    sum += arr[i];
}
\end{lstlisting}
\end{scriptsize}

\begin{scriptsize}
\begin{lstlisting}[language=c, label={do-all-ex}, caption=Do-all loop example]
#pragma omp parallel for
for (int i = 0; i < n; ++i) {
    arr[i] = i + 1;
}
\end{lstlisting}
\end{scriptsize}

\begin{scriptsize}
\begin{lstlisting}[language=c, label={private-ex}, caption=private loop example]
#pragma omp parallel for private(temp)
for (int i = 0; i < n; i++) {
    temp = array[i] * 2;  
    result[i] = temp;
}
\end{lstlisting}
\end{scriptsize}

\begin{scriptsize}
\begin{lstlisting}[language=c, label={private-ex}, caption=Combination of reduction and private loop example]
#pragma omp parallel for private(temp) reduction(+:sum)
for (int i = 0; i < n; i++) {
    temp = array[i] * 2;  
    sum += temp;
}
\end{lstlisting}
\end{scriptsize}

\subsection{Hardware Specifications}
\label{ref:hardware-spec}

The runtime experiments are performed on compute cluster with Slurm workload manager. Each compute node is invoked using a job script like below.

\begin{scriptsize}
\begin{lstlisting}[language=c, label={job-script}, caption=Basic job script for running the benchmarks]
#!/bin/bash
#SBATCH --nodes=1 
#SBATCH --cpus-per-task=1  
# max memory allocated for the job
#SBATCH --mem=8G 
# time allocated for the job
#SBATCH --time=0:10:00 
# output file name
#SBATCH --output=output.out 
# error file name
#SBATCH --error=error.err 
# name of the job
#SBATCH --job-name="running-job" 
# select CPU architecture
# (biocrunch = Intel Xeon Gold 6152, swift= AMD EPYC 7543P)
#SBATCH --partition=biocrunch/swift 

# -------- instructions for running benchmarks here --------
\end{lstlisting}
\end{scriptsize}

Table \ref{tbl:hardware-specifications} shows the detailed configuration of the CPU architectures that are used for the runtime experiments. 

\begin{table}[h]
\caption{Hardware Specifications}
\vspace{-17pt}
\label{tbl:hardware-specifications}
\begin{center}
    \setlength\tabcolsep{10pt}
    \resizebox{1\columnwidth}{!} {%
\begin{tabular}{ccc}
\multicolumn{1}{c}{\textbf{Features}} & \multicolumn{1}{c}{\textbf{Intel Xeon Gold 6152}} & \multicolumn{1}{c}{\textbf{AMD EPYC 7543P}} 
\\ \hline
Total Cores & 22 & 32\\
Total Threads & 44 & 64\\
Max Frequency & 3.7 GHz & 3.7 GHz\\ 
Processor Base Frequency & 2.1 GHz &  2.8 GHz\\ 
Cache & 30.25 MB L3 Cache & 256 MB L3 Cache\\
Memory Types & DDR4-2666 & DDR4\\
\end{tabular}
}
\vspace{-10pt}
    \end{center}
\end{table}

\subsection{Loss curves, training time and Hyparameters of GNN classifiers}
\label{sec:appendix-loss--training-times}

\textbf{Hyperparameters:} We experimented with different hidden layer sizes and learning rates and ultimately chose 64 as the hidden layer size and set the learning rate to 0.01. Each node of the heterogeneous \perfograph is embedded to a 120-dimensional vector. Therefore, the input layer size is set to 120. The output layer size is set to the number of classes, which is 2, as all three of the GNN models do binary classification. For graph-level prediction, the `mean' aggregation function combines the results of different node types, and finally linear classifier is used in the last layer of the RGCN model. The linear classifier produces a probability score for each class. The models are trained for 120 epochs, and the checkpoint with the highest validation accuracy is saved for later inference. 

\begin{figure}[]
  \centering
  \includegraphics[width=\linewidth]{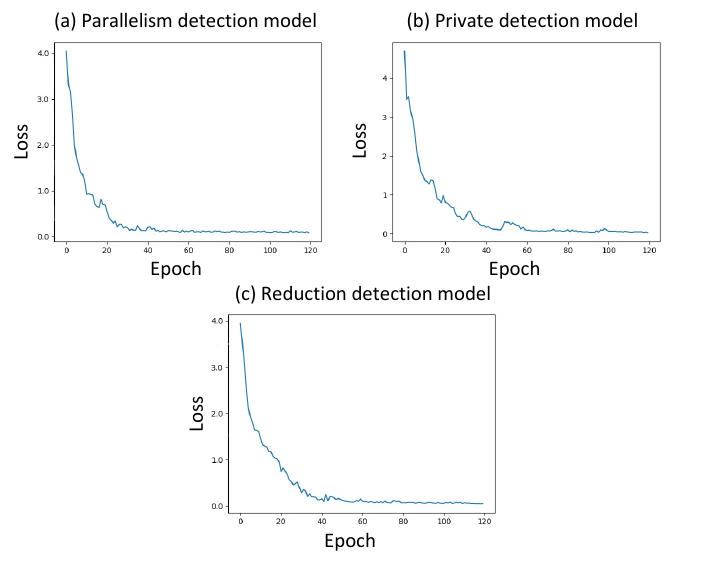}
  \vspace{-20pt}
  \caption{Epoch vs. loss curves for the three GNN-based parallelism and pragma detection modules}
  \label{fig:epoch-loss-curves}
  \vspace{-15pt}
\end{figure}

Table ~\ref{tbl:training-time} shows the training time required for all three models. Figure ~\ref{fig:epoch-loss-curves} shows the epoch vs. loss curve for all three GNN models. 
\vspace{-0.8em}

\begin{table}[h]
\caption{The training time required for the three GNN-based predictors}
\vspace{-17pt}
\label{tbl:training-time}
\begin{center}
    \setlength\tabcolsep{1pt}
    \resizebox{0.6\columnwidth}{!} {%
\begin{tabular}{cc}
\multicolumn{1}{c}{\textbf{Model}} & \multicolumn{1}{c}{\textbf{Training  Time}}
\\ \hline
Parallelism Detection model & 17 min 5 sec \\
Private Detection model & 9 min 27 sec \\
Reduction Detection model & 9 min 12 sec \\
\end{tabular}
}
\vspace{-10pt}
    \end{center}
\end{table}


\subsection{Ensuring Correctness of the generated codes}
\label{sec:appendix-correctness}

When the GNN predicts that a given loop is not parallel \ourtool \xspace{} does not generate any parallel code for that. In this case, \ourtool \xspace{} outputs the original sequential loop. Hence, correctness is preserved.  Like any other ML model, there will be False Positives (FP) and False Negatives (FN). Since all loops in the Rodinia Benchmark are correctly classified by \ourtool, we will explain how FP and FN are handled for the NAS benchmark. During our experiments with NAS benchmark, 30 out of 32 non-parallel loops are correctly detected by \ourtool. These 30 loops that are detected as non-parallel do not need to be checked further. Because we know that even if there are some parallel loops in this set, treating them as sequential loops will only hurt performance but not correctness. There are 90 loops in the test set of NAS benchmark. That means using \ourtool, we can filter out 33.33\% of loops for analysis safely. Hence, developers' workload is reduced by 33.33\%.

Only the remaining loops need to be analyzed further after applying \ourtool, as they may contain the following scenarios: 
\vspace{-2pt}
\begin{itemize}
    \item Non-parallel loop detected as parallel: Only 2 non-parallel loops are wrongly detected as parallel; we use the original sequential version of these two loops to maintain correctness during execution. 
    \item Wrong OpenMP clause prediction: Only four loops have been decorated with the wrong OpenMP clauses. We also use the sequential version of these 4 loops to maintain correctness. Note that using proper OpenMP clause will result in speedup, but that will give an unfair advantage to \ourtool \xspace{} while calculating performance gain during execution. Hence, we use the sequential version as \ourtool \xspace{} failed to detect the right clauses.
\end{itemize}

Therefore, only 6 out of the 60 analyzed loops required manual fixing. The rest 54 loops are already correctly generated by \ourtool. The outputs of the generated codes are compared with the original outputs to ensure that the generated codes produce results identical to the original codes.

\subsection{Structure of \perfograph \xspace{} \& Node and Edge Embedding Generation}
\label{sec:appendix-perfograph}

Each loop is first converted to IR, and then from that IR, we create the \perfograph \xspace{} representation of that loop. There are three types of nodes:
\vspace{-3pt}
\begin{itemize}
    \item Control nodes: Each control node represents a statement in IR. Tokens in the IR statement are considered as features for the control node. We embed each token of the statement and finally concatenate the embedding of all tokens to generate the final embedding for a control node.
    \item Variable nodes: \perfograph \xspace{} contains only the type of a variable in variable nodes. So, the type is considered as the feature for variable nodes, and embedding is generated for the type token.
    \item Constant nodes: For constant nodes, \perfograph \xspace{} representation contains both type and value, so we consider both of them as features. First, we generate the embedding for the type token. For generating the embedding for value-token, we use the Digit Embedding as described in \perfograph \xspace{} paper. Finally, the type-token and value-token embeddings are concatenated to generate the final embedding for each constant nodes.

\end{itemize}

The variable and constant nodes represent the variables and constants that are associated with those IR statements (control nodes). There are three types of edges: 

\begin{itemize}
    \item Control flow edges: Represents the flow of the program.
    \item Data flow edges: Represents the data dependencies of different nodes in the program.
    \item Call flow edges: Represents the functional call dependencies of the program.
\end{itemize}

Nodes are connected with each other using these three different edges. Edge embeddings are obtained using a one-hot-encoding approach. All the embeddings mentioned are generated using the default Pytorch learnable embedding mechanism.

\subsection{ParaBLEU implementation}
\label{sec:appendix-ParaBLEU}

We modified ParaBLEU score following the same idea of ~\cite{wen2022babeltower}. It is adjusted for OpenMP directives. The modified formula of the ParaBLEU score, which is suitable for OpenMP code evaluation, is shown below:

\vspace{0.2cm}

$ParaBLEU_{OMP}=\alpha*BLEU+\beta*BLEU_{OpenMPkeywords}$

\vspace{0.2cm}

In this formula, $\alpha$ and $\beta$ are the heuristic parameters that we set both of them as $0.5$ for our evaluation. These ratios specify the contribution of the original BLEU score and the weighted BLEU score, highlighting the similarities between n-grams containing our selected OpenMP keywords.

\subsection{Runtime Experiments - NAS Parallel Benchmark Input}
\label{sec:appendix-input-nas}

We execute the applications of NAS Parallel Benchmark suite using CLASS A input that comes along with the benchmark. Table \ref{tbl:nas-problem-size} shows the size of the problems and the number of iterations for each application for CLASS A. Detailed runtimes for each application can be found in the \href{https://github.com/quazirafi/AutoParLLM/tree/main/DetailedRuntimes}{repository link}.

\begin{table}[h]
\caption{Input problem size for each application in NAS Parallel Benchmark Suite}
\vspace{-17pt}
\label{tbl:nas-problem-size}
\begin{center}
    \setlength\tabcolsep{10pt}
    \resizebox{1\columnwidth}{!} {%
\begin{tabular}{ccc}
\multicolumn{1}{c}{\textbf{Applications}} & \multicolumn{1}{c}{\textbf{Problem Size}} & \multicolumn{1}{c}{\textbf{Number of iterations}} 
\\ \hline
IS & $2^{23}$ & 10 \\
EP & $2^{28}$ & 1  \\ 
FT & 256 × 256 × 126 & 6  \\
CG & 14000 & 15 \\
MG & 256 × 256 × 256 & 4  \\
LU & 64 × 64 × 64 & 250 \\
BT & 64 × 64 × 64 & 200  \\
SP & 64 × 64 × 64 & 400  \\
\end{tabular}
}
\vspace{-10pt}
    \end{center}
\end{table}

\subsection{Runtime Experiments - Rodinia-3.1 Benchmark Input}
\label{sec:appendix-input-rodinia}

Rodinia-3.1 Benchmark applications are also executed using the inputs that come along with the benchmark itself. Below we provide details regarding the inputs for running the applications.

\begin{itemize}
    \item BFS is executed with the input `graph1MW\_6.txt', which applies BFS on a graph that contains 1000000 nodes and 5999970 edges.
    \item B+Tree is executed with the inputs `mil.txt' and `command.txt' with the block size (6000, 3000) and key size of 10000.
    \item Heartwall is executed with the input `test.avi' which contains a series of medical ultrasound images with a framesize of 10.
    \item 3D is executed with a 3D grid size (512, 8, 100) along with inputs power\_512x8 and temp\_512x8, where the last two inputs represent power dissipation and initial temperature at each grid-point for thermal simulations.
\end{itemize}

We provide detailed runtimes for each application in Rodinia in the \href{https://github.com/quazirafi/AutoParLLM/tree/main/DetailedRuntimes}{repository link}.

\subsection{Runtime Experiments - Compare Average Speedup Across DIfferent LLMs}
\label{appendix:speeup-all-llms}

\begin{figure*}[h]
    \centering
    \includegraphics[width=1\linewidth]{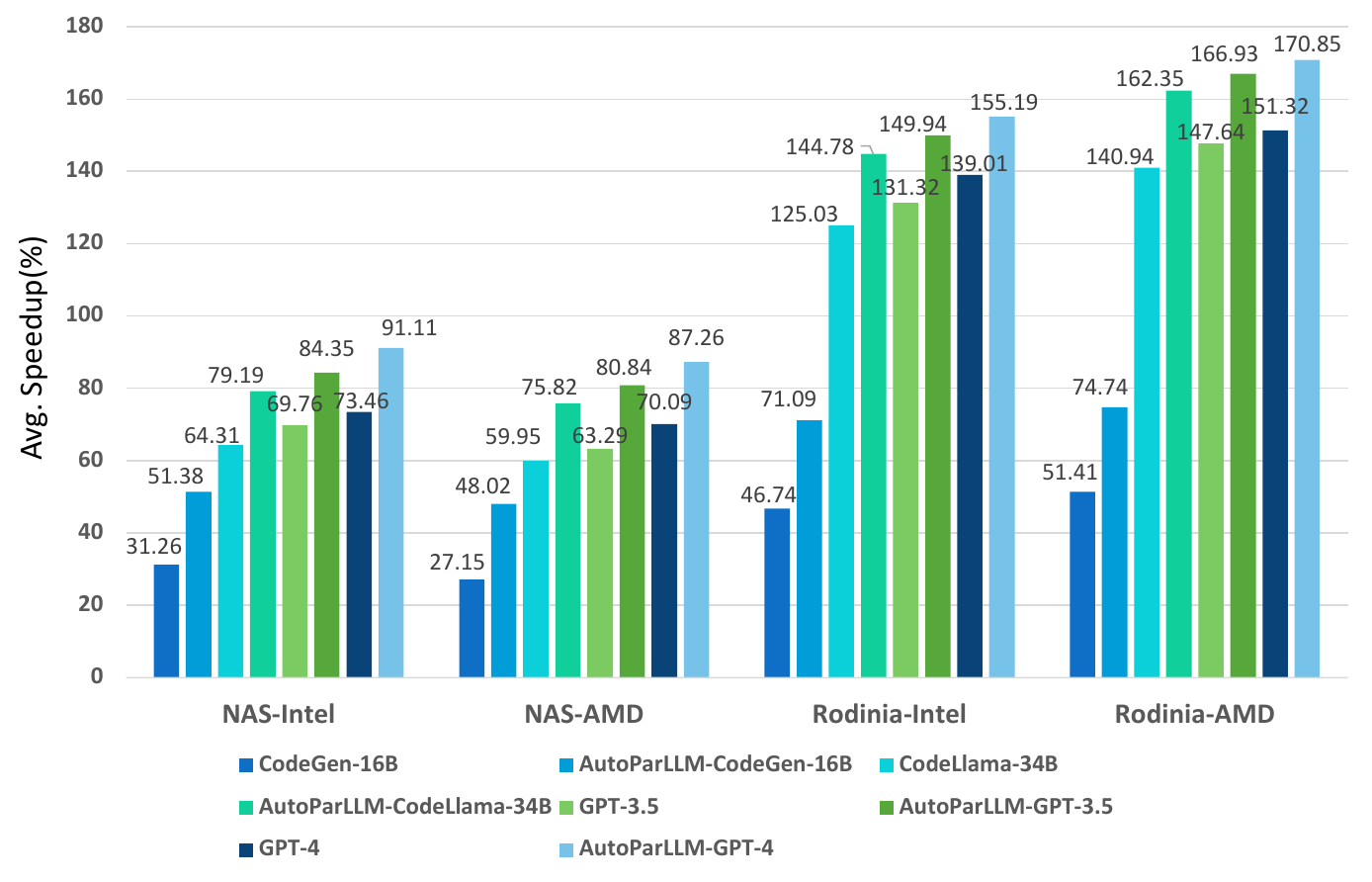}
    \vspace{-25pt}
    \caption{Effect of AutoParLLM on different LLMs. LLMs are prompted with few shot settings \& speedups are reported using 4 threads.}
    \vspace{-10pt}
    \label{fig:summary-speeup-all-llms}
\end{figure*}

Figure \ref{fig:summary-speeup-all-llms} shows the effect of \ourtool \xspace on four LLMs CodeGen, CodeLlama, GPT-3.5 and GPT-4. \ourtool \xspace improves the average speedup gain of all four LLMs on both NAS and Rodinia Benchmark.

\subsection{Runtime Experiments - Different Thread Configurations}
\label{sec:appendix-thread-config}

\begin{figure*}[]
  \centering
  \includegraphics[width=\linewidth]{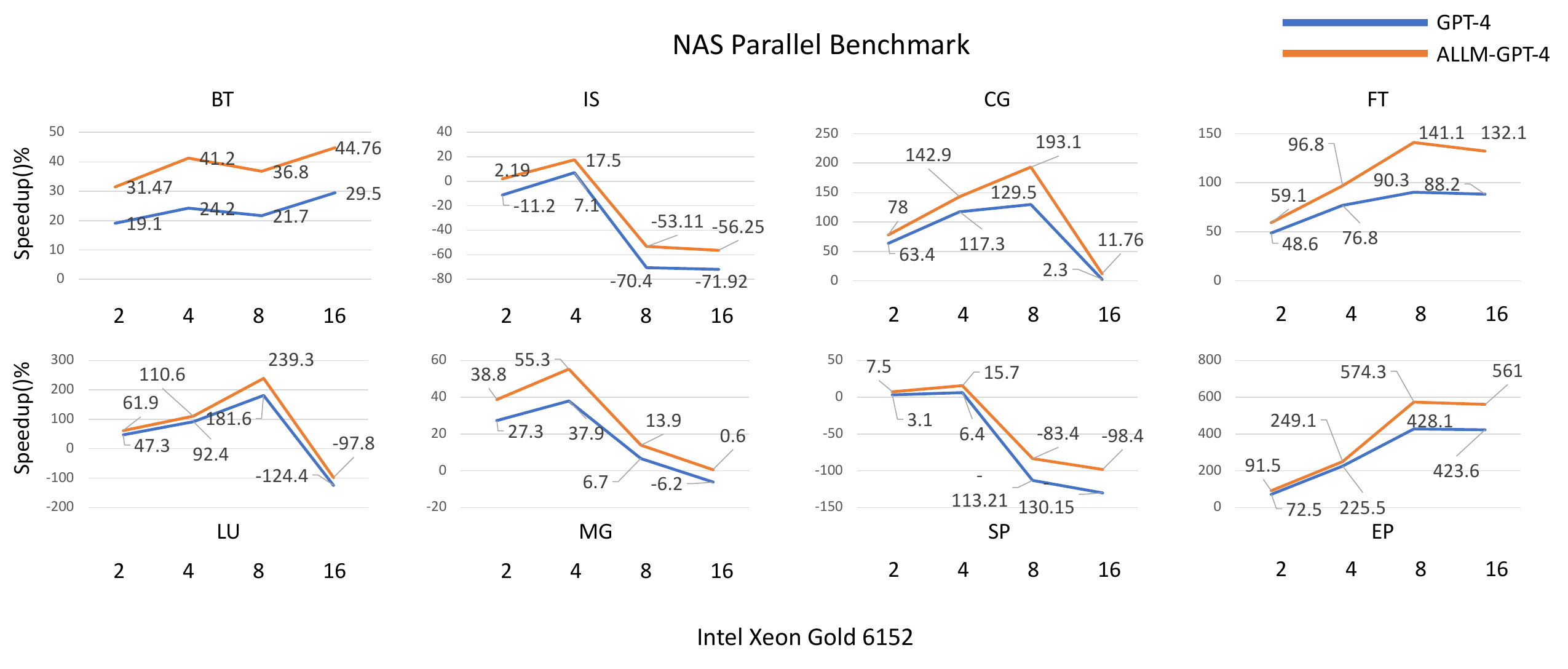}
  \vspace{-20pt}
  \caption{Comparing Effects of AutoParLLM on GPT-4 for different thread configurations for NAS Benchmark on Intel Xeon Gold Machine.}
  \label{fig:nas-thread-intel}
  \vspace{-15pt}
\end{figure*}

\begin{figure*}[]
  \centering
  \includegraphics[width=\linewidth]{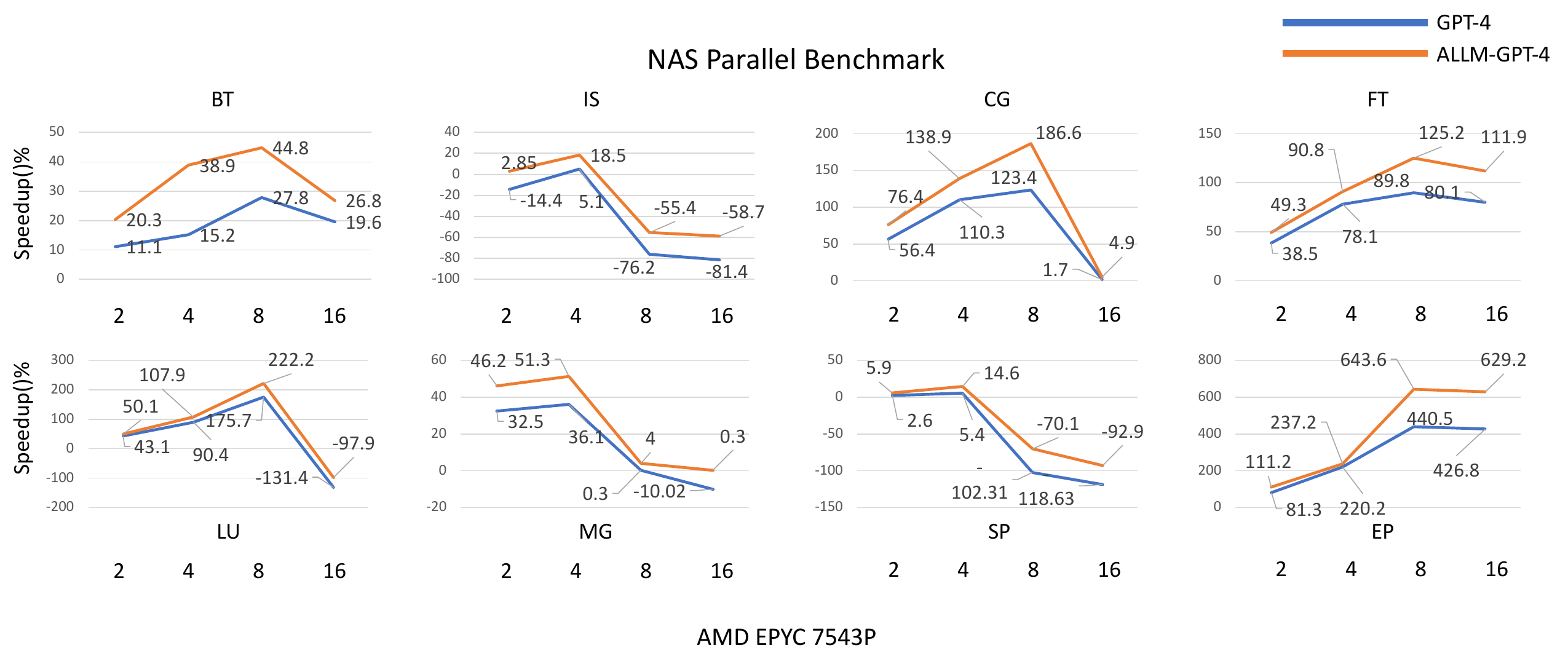}
  \vspace{-20pt}
  \caption{Comparing Effects of AutoParLLM on GPT-4 for different thread configurations for NAS Benchmark on AMD EPYC Machine.}
  \label{fig:nas-thread-amd}
  \vspace{-15pt}
\end{figure*}

We analyzed the effect of \ourtool on GPT-4 with varying thread numbers. GPT-4 is chosen as, according to our study, it performed the best among the other LLMs. We experimented with 4 thread configurations: 2, 4, 8 and 16. From the results of Figure \ref{fig:nas-thread-intel}, \ref{fig:nas-thread-amd} and \ref{fig:rod-thread-all}, it can be observed that in all configurations, the speedup up obtained by ALLM-GPT-4 is better than the base GPT-4.

  \label{fig:rod-thread-all}

\begin{figure*}[]
  \centering
  \includegraphics[width=\linewidth]{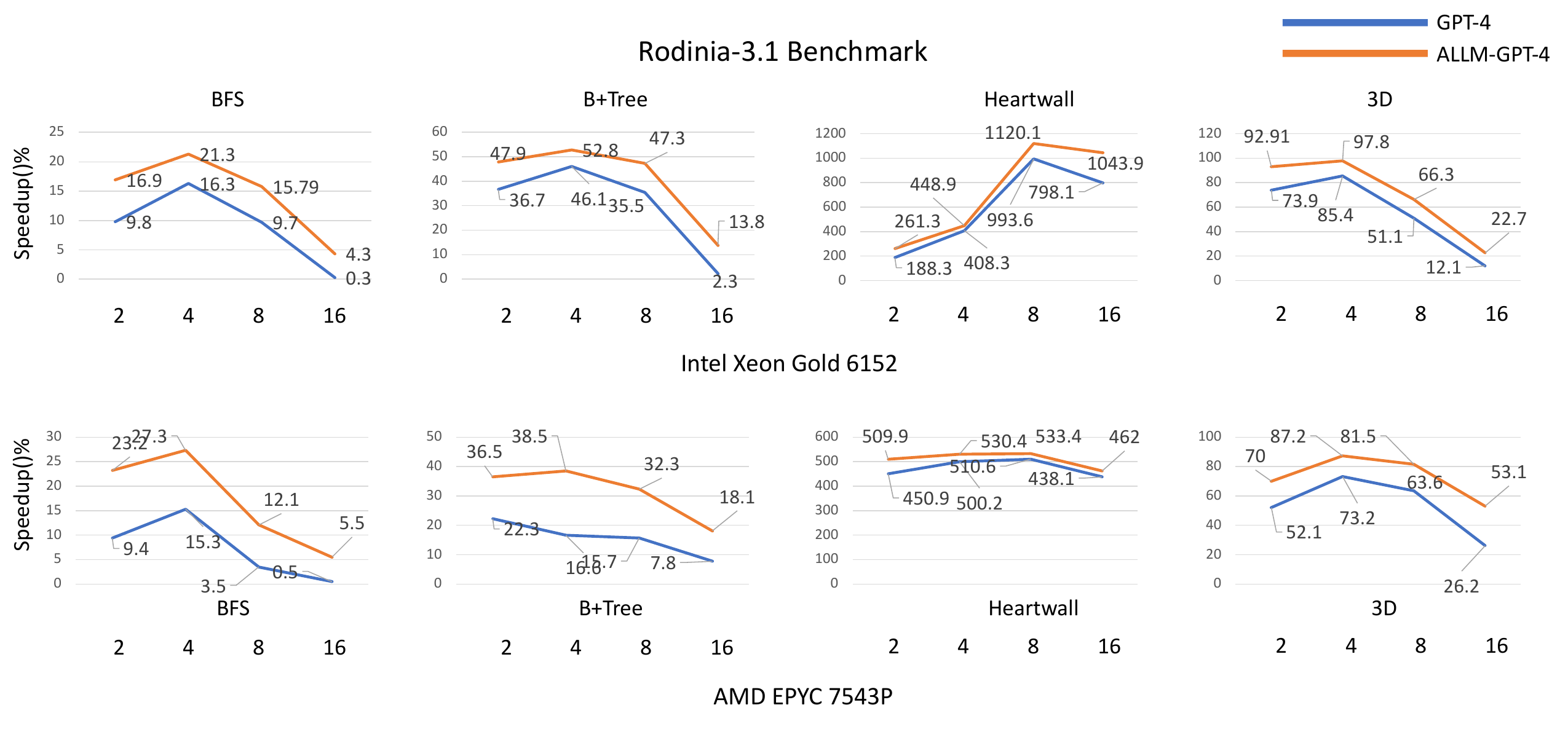}
  \vspace{-20pt}
  \caption{Comparing Effects of AutoParLLM on GPT-4 for different thread configurations for Rodinia-3.1 Benchmark on both Intel Xeon Gold and AMD EPYC Machine}
  \label{fig:rod-thread-all}
  \vspace{-15pt}
\end{figure*}

\subsection{Traditional Approaches Comparison}
\label{sec:appendix-traditional}

In Table ~\ref{discopop-table}, we compare the parallelism discovery of \ourtool \xspace{} against three popular parallelization tools: Pluto, AutoPar, and DiscoPoP on the DiscoPoP subset (1226 files) of OMP\_Serial Dataset. The task is to detect whether a code can be parallelized or not. There are also two other subsets: Pluto Subset and AutoPar Subset in the OMP\_Serial dataset. However, DiscoPoP can not process some of the codes in those two subsets as it requires each code to be executable for proper analysis. So, to have a fair comparison with all three tools, we choose the DiscoPoP subset for this experiment as all codes in the DiscoPoP subset are executable. Pluto and AutoPar perform static analysis to find parallel regions (such as loops), whereas DiscoPoP is a dynamic analysis tool that executes the code to identify potential parallel regions. Therefore, choosing these three tools enables us to compare \ourtool \xspace{} against both static and dynamic analysis tools. Due to program execution, from Table ~\ref{discopop-table}, we see that DiscoPoP gains an advantage over Pluto and AutoPar as it has 15\% and 8\% better accuracy, respectively. However, it can be observed that \ourtool \xspace has far better accuracy than all three tools.
\vspace{-9pt}
\begin{table}[h]
\caption{Parallelism Discovery on DiscoPoP Subset (Detecting parallel loops)}
\vspace{-10pt}
\label{discopop-table}
\begin{center}
\resizebox{0.9\columnwidth}{!} {%
\begin{tabular}{lllll}
\multicolumn{1}{c}{\bf Tool}  &\multicolumn{1}{c}{\bf Precision} &\multicolumn{1}{c}{\bf Recall} &\multicolumn{1}{c}{\bf F1-score} &\multicolumn{1}{c}{\bf Accuracy} 
\\ \hline \\
Pluto   &    \bf 1   &   0.38   &   0.55  & 0.48 \\
AutoPar   &    \bf 1   &   0.49   &   0.67  & 0.55 \\
DiscoPoP   &    \bf 1   &   0.54   &   0.70  & 0.63 \\
\textbf{\ourtool}   &    0.99   &  \bf 0.99   &  \bf 0.99  & \bf  0.99 
\vspace{-27pt}
\end{tabular}
}
\end{center}
\end{table}

\subsection{Comparing with Deep Learning Approaches}
\label{appendix:compare-deep-learning}
Here, we applied \ourtool \xspace on the Pluto, AutoPar, and DiscoPoP subsets of the OMP\_Serial Dataset, which consists of parallel and non-parallel codes. The task is the same as the previous. The Pluto subset has 4032 files, the AutoPar subset has 3356 files, and the DiscoPoP subset has 1226 files. The results of the tools Graph2Par ~\cite{chen2023learning}, PrograML and \perfograph \xspace are reported from \cite{tehranijamsaz2023perfograph}. We apply our \ourtool \xspace model on the same subsets and compare these approaches. Table \ref{tab:par-1} shows that \ourtool \xspace surpasses the performance of the state-of-the-art \perfograph \xspace \cite{tehranijamsaz2023perfograph} by achieving as high as 6\% better accuracy.
\vspace{-5pt}
\begin{table}[!ht]

\captionsetup{justification=centering}
\caption{Performance comparison of \ourtool \xspace on parallelism discovery task in the OMP\_Serial dataset with existing approaches.}
\small
\setlength\tabcolsep{3.5pt}
\centering
\resizebox{0.9\columnwidth}{!} {%
\begin{tabular}{cccccc}
\hline
Subset           & Approach & Precision & Recall & F1-score & Accuracy \\ \hline
\multirow{3}{*}{Pluto} 
                          & Graph2par         & 0.88               & 0.93            & 0.91              & 0.86              \\  
                          & \programl          & 0.88               & 0.88            & 0.87              & 0.89              \\  
                          & \perfograph  & 0.91               & 0.90            & 0.89                     & 0.91     \\ 
                          & \textbf{\ourtool}  & \bf 0.97               & \bf 0.98            & \bf 0.98              & \textbf{0.96}     \\ \hline
\multirow{3}{*}{AutoPar}
                          & Graph2par         & 0.90               & 0.79            & 0.84              & 0.80              \\  
                          & \programl          & 0.92               & 0.69            & 0.67              & 0.84              \\  
                          & \perfograph  & 0.85               & 0.91            & 0.85                    & 0.86     \\ 
                          & \textbf{\ourtool}  & \bf 0.93               & \bf 0.92            & \bf 0.93              & \textbf{0.92}     \\ \hline
\multirow{3}{*}{DiscoPoP}
                          & Graph2par         & 0.90               & 0.79            & 0.84              & 0.81              \\  
                          & \programl          & 0.92               & 0.94            & 0.92              & 0.91              \\  
                          & \perfograph &  0.99               & \bf 1               &  0.99                     &  0.99     \\ 
                          & \textbf{\ourtool} & \bf 0.99               & 0.99            & \bf 0.99              & \textbf{0.99} 
    \vspace{-20pt}
\end{tabular}
}
\label{tab:par-1}
\end{table}

\subsection{Analyzing Wrong Predictions in OpenACC}
\label{sec:appendix-openacc-mismatch}

\begin{figure}[h]
    \includegraphics[width=1\linewidth]{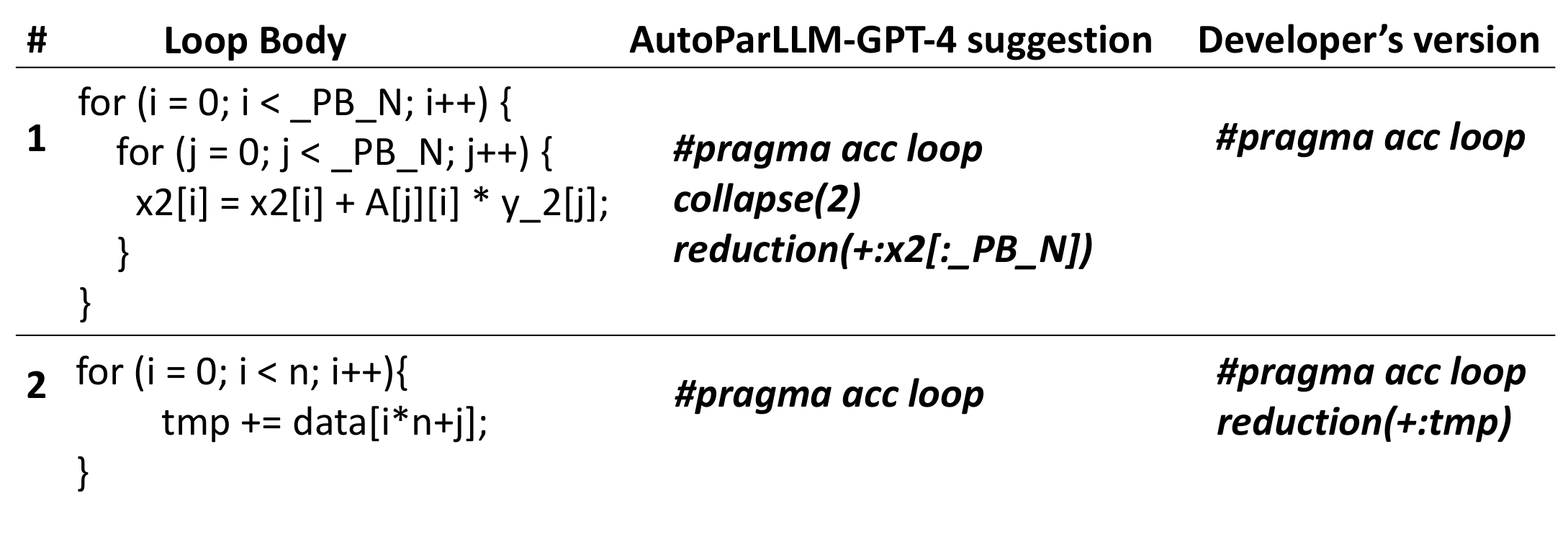}
    \vspace{-20pt}
    \caption{Mismatches of \ourtool \xspace in the OpenACC dataset.}
    \label{fig:openacc-mismatch}
\vspace{-12pt}
\end{figure}

Here we describe the wrong predictions in our OpenACC experiment.  The test set contains 12 parallel (5 \texttt{private}, 7 \texttt{reduction}) and 5 \texttt{non-parallel} loops. For generating the complete OpenACC clauses we use the predicted patterns from \ourtool \xspace and incorporate them into the prompts for the GPT-4. However, instead of OpenMP now we instruct the \ourtool-GPT-4 to parallelize the code using OpenACC. There were 2 wrong predictions by \ourtool-GPT-4. For the first case, in Figure \ref{fig:openacc-mismatch}, the developer's implementation from the benchmark only parallelizes the outermost loop. However, \ourtool-GPT-4 configuration suggested two extra clauses: \texttt{collapse} and \texttt{reduction}. \texttt{collapse} merges the two loops in a flattened 2D iteration space, allowing OpenACC to parallelize both loops. \texttt{reduction} ensures proper summation among threads and prevents any race conditions. Although this configuration does not match with the developer's version from benchmark it is more efficient as it parallelizes both of the loops. This shows the potential of \ourtool \xspace in finding optimization opportunities that may be overlooked by human developers. For the second case, \ourtool \xspace correctly detects it as a parallel loop however, it fails to identify the \texttt{reduction} operation. We hypothesize the reason is because of the presence of complex indexing on variable \texttt{data}.

\subsection{Detailed Related Works}
\label{sec:appendix-related-works}

Here we describe some of the existing approaches to automatic parallelization and also the current metrics for translation evaluation.

\subsubsection{Traditional Parallelism Assistant Tools} 

Traditional tools are based on static and dynamic analysis to automatically parallelize sequentially written programs. PLuTo \cite{bondhugula2008pluto} analyzes code statically and can optimize programs for parallel execution. It is a polyhedral source code optimizer based on OpenMP \cite{meadows2007openmp}. Rose \cite{quinlan2011rose} is another static source-to-source compiler infrastructure that also supports automatic parallelization using OpenMP. Both of these tools do not require or use code runtime information while generating parallel counterparts of sequential programs. DiscoPoP \cite{li2016unveiling} is a dynamic analysis-based parallelism assistant tool. It uses dynamic control-flow analysis and data-dependence profiling for identifying parallel regions in source programs. The works of \cite{huda2015using} use the output of DiscoPoP to find parallel patterns using template-matching techniques. The DiscoPoP-generated hybrid dependence analysis results have been used by \cite{huda2016automatic} to detect various parallel patterns like Geometric decomposition, reduction and task-based parallelism. However, these traditional analysis-based tools have some drawbacks, as pointed out by \cite{chen2023learning}, and they miss a lot of parallelism opportunities due to being overly conservative.

\subsubsection{Data-driven Approaches}

With the significant progress in the field of machine learning and deep learning, many have proposed automatic data-driven approaches to identify parallelism opportunities and suggest appropriate constructs. Authors of 
 \cite{chen2023learning} proposed an approach based on graph neural networks and augmented Abstract Syntax Trees (ASTs) to identify parallel loops. Their results show that their GNN-based approach outperforms PragFormer \cite{harellearning}, which uses Transformers  \cite{vaswani2017attention} to discover parallelism opportunities in code.
 \cite{shen2021towards} uses contextual flow graphs with graph convolution neural networks~\cite{kipf2016semi} to detect parallelism.  \cite{shen2023multigraph} uses a combination of control flow, data flow, and abstract syntax tree to predict parallelism. There are also attempts to use IR-based representation for automatic parallelization of ML models \cite{schaarschmidt2021automap}.
Even though there has been some progress in adapting machine learning techniques to predict parallelism opportunities, little effort has been applied to connect parallelism detection and code generation. In this work, we address this gap. We connect GNNs and LLMs to not only discover parallelism but also generate parallel code.
Connecting GNNs to LLMs has been investigated recently \cite{ghosh2022graph,chen2023exploring}. However, to the best of our knowledge, we are the first to leverage the result of GNN to guide LLM to generate parallel code out of serial code.

\subsubsection{LLM-based Approaches}

Large Language Models (LLMs) have achieved remarkable success across various domains, including parallel code generation. \citet{chen2023lm4hpc} were pioneers in applying LLMs to high-performance computing (HPC) tasks, including parallelism detection. They compared the performance of GPT-3.5 with their previous GNN-based approaches~\citep{chen2022multi, chen2023learning}. Their findings demonstrated that GPT-3.5 could achieve competitive performance in parallelism detection with even basic prompts. In subsequent research, they trained an LLM specifically for OpenMP pragma generation, introducing a tailored chain-of-thought approach~\citep{chen2024ompgpt}.  \citet{nichols2024can} evaluated the performance of the parallel code generated by LLMs and indicated that even with great potential, noting that while LLMs show great potential, there remains a significant gap between the generated code and the expectations of the HPC community. They also explored fine-tuning LLMs for generating optimized parallel code~\citep{nichols2024hpc, nichols2024performance}.
In contrast to these prior works, our approach uniquely combines the strengths of GNNs with LLMs to generate prompts enriched with external knowledge. This hybrid method leverages the advantages of both GNNs and LLMs, providing a versatile solution compatible with any LLM.

\subsubsection{Metrics for Translation Evaluation} 

The BLEU score~\cite{2002PapineriBLEU} is a classical metric for evaluating textual similarity in machine translation. It assesses the overlap between sequences of consecutive $n$ words, called n-grams.
Meteor ~\cite{banerjee2005meteor} was introduced to address some of the limitations of the BLEU score, such as its tendency to underestimate high-order n-grams.
ROUGE ~\cite{lin2004rouge} encompasses a set of metrics that evaluate various aspects of textual similarity.
To evaluate code generation, recent works have introduced metrics like CodeBLEU ~\cite{ren2020codebleu} and CodeBERTScore ~\cite{zhou2023codebertscore}. Authors of BabelTower ~\cite{wen2022babeltower} designed a metric called "ParaBLEU" specifically for evaluating parallel codes. However, the designed metric is limited to evaluating CUDA code only.
So, none of these metrics have been specifically designed for evaluating the quality of OpenMP constructs in terms of textual similarity.

\subsection{Experimenting with Randomized Few Shot COT}\label{appendix:randomized_few_shot_prompting}

Here, we discuss the results of our experiments regarding adding randomly sampled examples in GPT-4 prompts. Instead of handpicking samples from different parallelization scenarios, here, the samples for constructing the prompt are randomly picked from the dataset. To enable consistent comparison with the hand-picked sampling technique we also limit the sample number to 5 samples. The results are presented for both NPB and Rodinia benchmark in Table \ref{npb-rand-prompt-scores} and Table \ref{rod-rand-prompt-scores}, respectively. From the results, it can be observed that the performance and of hand-picked sample based prompting (Man) and randomized sampling based (Rand) techniques yields very close results in most cases. However, \ourtool \xspace surpasses both prompting techniques by a significant margin in both benchmarks in terms of all three metrics scores.

\begin{table}[!ht]
    \caption{Results on NPB Benchmark Suite for Randomized Few Shot Prompting}
    \vspace{-12pt}
    \label{npb-rand-prompt-scores}
    \begin{center}
    \setlength\tabcolsep{2pt}
    \resizebox{1\columnwidth}{!} {%
    \begin{tabular}{ccccccccc}
    \multicolumn{1}{c}{\bf Model}  &\multicolumn{1}{c}{\bf CBTScore} &\multicolumn{1}{c}{\bf ParaBLEU} &\multicolumn{1}{c}{\bf OMPScore} \\ \hline

    0-shot-COT-GPT-4            & 73.8 & 27.49  & 46.4\\
    few-shot-COT-GPT-4 (Rand)   & 76.3 & 35.69  & 58.64\\
    few-shot-COT-GPT-4 (Man)    & 76.5 & 29.97  & 58.06\\
    \bf \ourtool-GPT-4          & \bf 96.4 &  \bf 48.48  & \bf 95.15 \\
    \end{tabular}
    }
    \end{center}
\end{table}

\begin{table}[!ht]
    \caption{Results on Rodinia-3.1 Benchmark Suite for Randomized Few Shot Prompting}
    \vspace{-12pt}
    \label{rod-rand-prompt-scores}
    \begin{center}
    \setlength\tabcolsep{2pt}
    \resizebox{1\columnwidth}{!} {%
    \begin{tabular}{ccccccccc}
    \multicolumn{1}{c}{\bf Model}  &\multicolumn{1}{c}{\bf CBTScore} &\multicolumn{1}{c}{\bf ParaBLEU} &\multicolumn{1}{c}{\bf OMPScore} \\ \hline

    0-shot-COT-GPT-4           & 91.2 & 49.14 & 79.37\\
    few-shot-COT-GPT-4 (Rand)  & 93.8    & 55.67  & 89.43  \\
    few-shot-COT-GPT-4 (Man)   & 92.12 & 54.43 & 89.01\\
    \bf \ourtool-GPT-4         & \bf 98.6 & \bf 69.09  & \bf 98.1 \\

    \end{tabular}
    }
    \end{center}
\end{table}

\clearpage

\subsection{Few Shot COT Prompt Example}
\label{sec:appendix-few-shot}

\begin{verbatim}

Qs: Parallelize the following code using OpenMP
for (int i = 0; i < n; ++i) {
arr[i] = arr[i-1] + arr[i+1];
}
Ans.
1. The loop contains inter-iteration dependencies at i-1 and i+1.
2. Hence loop can not be parallelized.

Qs: Parallelize the following code using OpenMP
for (int i = 0; i < n; ++i) {
sum += arr[i];
}
Ans.
1. The following code combines multiple iterations into a final outcome 'sum'
2. Hence adding reduction clause is necessary to parallelize
3. reduction should be added on 'sum' with '+' operator.
Qs: Parallelize the following code using OpenMP
for (int i = 0; i < n; ++i) {
arr[i] = i + 1;
}
Ans:
1. The code contains do-all pattern as all iterations are independent as no need 
to make any variable private to each thread.
2. Simply adding the 'parallel for' clause should be sufficient.
Qs:
for (int i = 0; i < n; i++) {
        temp = array[i] * 2;  
        result[i] = temp;
    }
Ans:
1. The code contains do-all pattern as all iterations are independent
2. But variable 'temp' needs to be private to each thread.
3. private clause should be added on 'temp'.
Qs:
for (int i = 0; i < n; i++) {
        temp = array[i] * 2;  
        sum += temp;
    }
Ans:
1. The code contains do-all pattern as all iterations are independent
2. But variable 'temp' needs to be private to each thread.
3. private clause should be added on 'temp'.
4. The following code combines multiple iterations into a final outcome 'sum'
5. Hence adding reduction clause is necessary to parallelize
6. reduction should be added on 'sum' with '+' operator.

Now Parallelize the following code using OpenMP:
<<input sequential code here>>

\end{verbatim}

\clearpage

\subsection{Zero Shot COT Prompt Example}
\label{sec:appendix-zero-shot}

\begin{verbatim}
Parallelize code using OpenMP by following the below rules: 
1. If loop iterations are independent of each other and no variable
is required to be private to each thread, then simply add 'parallel for'
clause
2. If loop iterations are independent but there is variable that needs
to be private to each thread then apply privatization through 
'private' clause
3. If a variable combines results from multiple loop iterations and
finally computes the output then apply reduction on the variable 
along with the associative operation.
4. It is possible that combinations of the above cases may arise.
<<input sequential code>>
\end{verbatim}

\end{document}

%% file: sections/intro.tex
\section{Introduction}
\begin{figure}[h]
    \centering
    \includegraphics[width=1\linewidth]{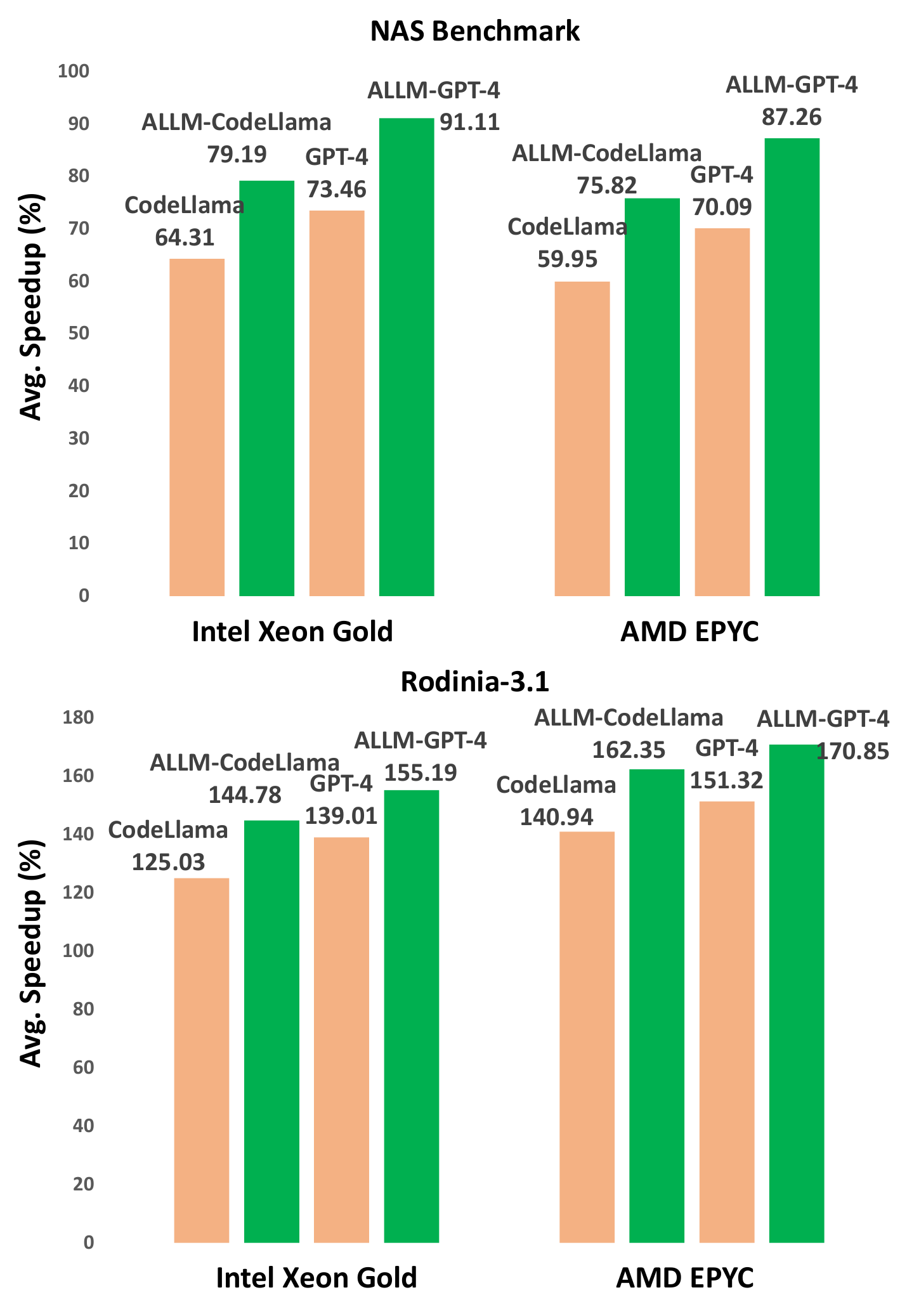}
    \vspace{-25pt}
    \caption{Effect of AutoParLLM. \textbf{ALLM = AutoParLLM applied (Green Bars).} Average speedup(\%) gain of GPT-4 is improved by \textbf{17.7\%} (Intel) \& \textbf{17.2\%} (AMD) on NAS and by \textbf{16.1\%} (Intel) \& \textbf{19.5\%} (AMD) on Rodinia. LLMs are prompted with few shot settings \& speedups are reported using 4 threads. (Comparison with more LLMs in Appendix \ref{appendix:speeup-all-llms}.)}  
    \vspace{-25pt}
    \label{fig:summary-speeup-nas}
\end{figure}


The rise in the number of on-chip cores has led to more frequent development of parallel code~\cite{moore}.
Nevertheless, to unleash the capabilities of multi-core systems, the need for developing parallel programs will continue to grow.
However, developing parallel programs is not a trivial task. The communication among cores, effective data sharing among the threads, synchronization, and many other factors need to be considered while crafting parallel programs, which makes the process of developing parallel programs far more complex than serial ones.

HPC communities have published different tools and programming models to ease the process of moving from serial to parallel code.
One of the well-established parallel programming models is OpenMP~\cite{mattson1999openmp}, which is a directive-based programming model that allows users to parallelize sequential code with minimal changes. Most modern compilers recognize and support parallelization through OpenMP constructs.
However, even with OpenMP, developers must carefully decide which clauses or directives they need to use. Inappropriate usage of clauses can cause concurrency bugs such as data race or decrease performance. Recently,  Large Language Models have also been used to generate parallel codes. One of the recent works ~\cite{nichols2024can} showed that LLMs struggle to generate correct parallel codes using basic prompts. ICL can help LLMs to generate better results. ICL has been applied to both the training ~\cite{ wei2023symbol, gu2023pre}  and inference stages ~\cite{li2023finding, wang2024large, li2023mot, xu2023small, wei2022chain}. In all these works, context means providing the model with some sample input and the expected response before providing the test inputs. That involves constructing samples that are close to the test inputs or constructing enough samples such that all cases are covered. In practice, it may be difficult to construct such a set of sample inputs along with the expected outputs. This is especially true for code parallelization, as the length of the codes that are given as context can easily exceed the context length. To overcome this, we propose \ourtool \xspace{}, a framework that works at the specific inputs provided to LLMs and generates context that is specific to that input only.  \ourtool \xspace{} has two main components. Firstly, a GNN-based context generator uses GNN to model flow-aware dependencies, i.e., data, control, and call flow and generates relevant context regarding whether the code is parallelizable and what parallel configurations are suitable.
Secondly, the LLM-based code generator uses the context generated by the GNNs to create an enhanced prompt and then generates the parallel code based on the prompts that contain the related context. In this work, we focus on OpenMP-based parallelization. Due to the specific nature of OpenMP directives, where the order might be important in some cases, traditional code synthesis metrics may not be suitable to evaluate the quality of the generated OpenMP code. As such, we propose a new metric called \ourscore \xspace{} that is more suitable to measure the quality of generated OpenMP constructs.

In summary, our paper provides the following key contributions: 
\vspace{-5pt}
\begin{itemize}
    \item A novel approach, called \ourtool, leveraging GNNs to generate ``context" for automatic code parallelization using large language models. To the best of our knowledge, \ourtool \xspace{} is the first tool that generates context based on GNNs and then uses LLMs for generating parallel codes.
    \vspace{-5pt}
    \item Evaluation of \ourtool \xspace on well-established benchmarks such as NAS Parallel and Rodinia benchmarks. We evaluate in terms of CodeBERTScore, and speedup gain and also compare our GNN-based prompting with zero-shot-COT and few-shot-COT approaches.
    \vspace{-5pt}
    \item A new evaluation metric called \ourscore \xspace{} to assess the quality of generated OpenMP code.
    \vspace{-5pt}
\end{itemize}
This paper is organized as follows. In the next Section, we discuss some background regarding OpenMP based parallelization. Followed by Section 3, where our approach is explained in detail along with the OMPScore. Section 4 presents the experimental results. Section 5 describes some of the related works. Finally, Section 6 concludes the paper.

\vspace{-3pt}

%% file: sections/motivation.tex
\section{Background}\label{sec:background}
\vspace{-3pt}




\begin{figure*}[]
\centering
\begin{subfigure}[b]{\textwidth}
\centering
   \includegraphics[width=1\linewidth]{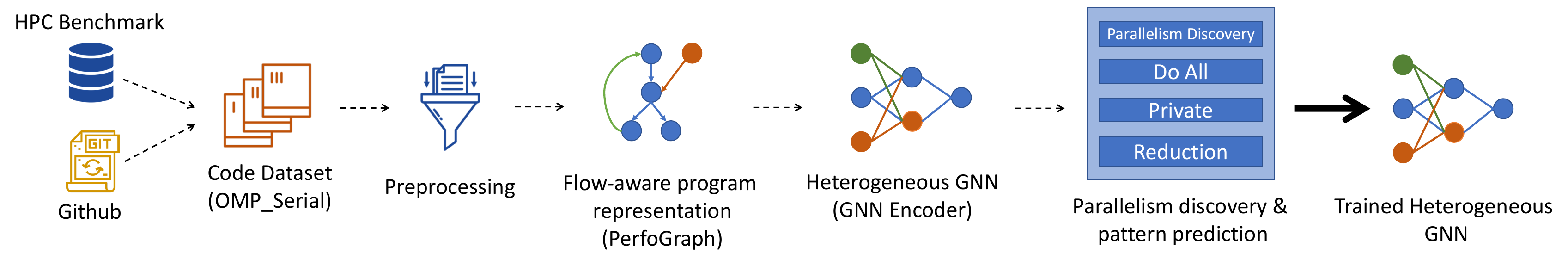}
   \caption{Training}
\end{subfigure}

\begin{subfigure}[b]{\textwidth}
\centering
   \includegraphics[width=1\linewidth]{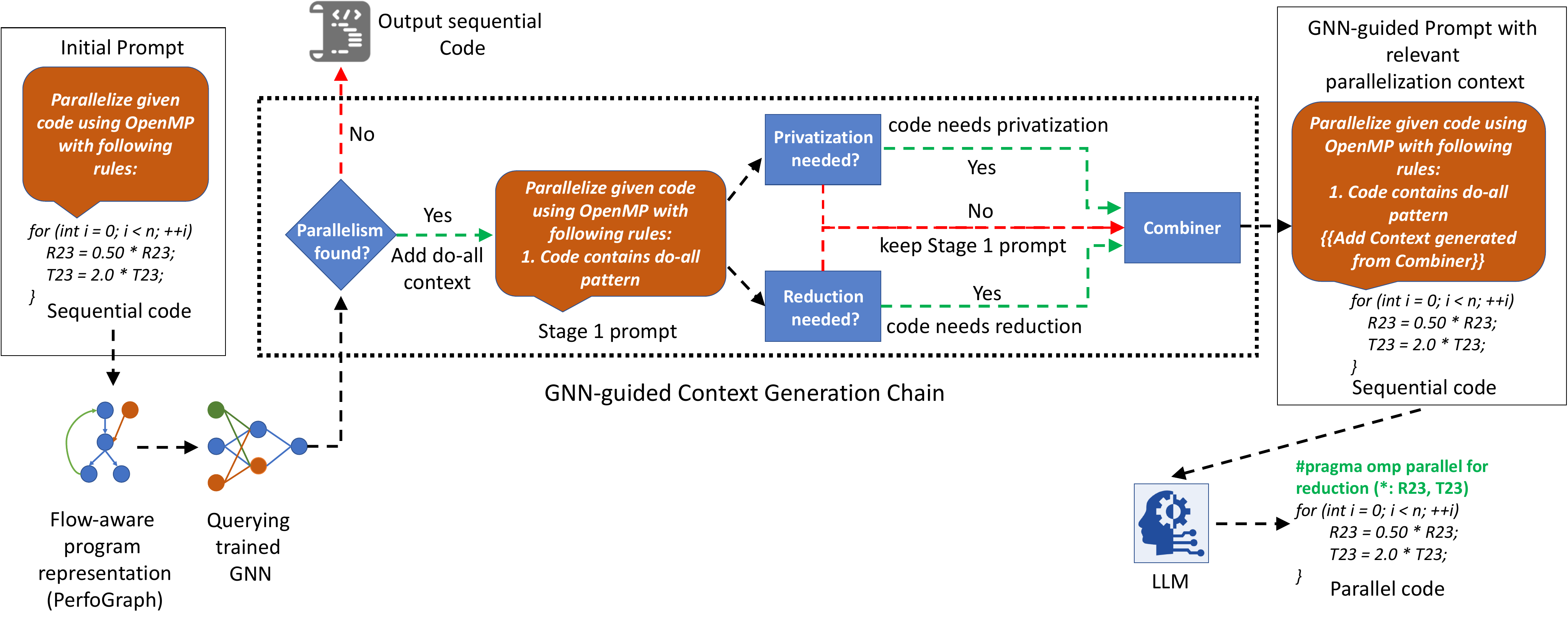}
   \caption{Inference}
\end{subfigure}
\vspace{-20pt}
\caption{Overview of the \ourtool \xspace{} workflow.}
\label{fig:workflow}
\vspace{-10pt}
\end{figure*}



OpenMP offers different types of parallelization configurations. This work focuses on loop-level parallelism. Not every configuration is applicable to every loop. For example, a loop having no inter-iteration dependencies can be parallelized by simply adding the \texttt{`\#pragma omp parallel for'} directive if there is no variable inside the loop context that needs to be \texttt{private} to each thread when parallel execution occurs. This type of loops are considered as \texttt{do-all}. However, there may be cases when a variable needs to be \texttt{private} to each thread such that any other thread running parallel can not modify the content of that variable, as it may result in inconsistency. Such cases are usually handled by privatization of the variable using the OpenMP clause \texttt{private}. Also, there may be cases where the result of multiple loop iterations is combined into a final output. This is usually known as a \texttt{reduction} operation. Such loops can also be parallelized by using \texttt{reduction} clause in OpenMP. It makes a separate copy of the \texttt{reduction} variable for each thread. When all threads are done with their calculations, it combines all the outputs of each separate copies of the variable and generates the final result. We provide examples of each in Appendix ~\ref{sec:appendix-background}.

%% file: sections/approach.tex
\section{Approach}
\vspace{-5pt}
In this section, we present \ourtool. A framework that leverages Graph Neural Network to learn the flow-aware characteristics of the programs, such as control flow, data flow, and call flow, to add additional context and guide LLMs to generate parallel code by constructing a GNN-guided OMP prompt.
Figure ~\ref{fig:workflow} shows the overall workflow of \ourtool.
In Figure ~\ref{fig:workflow}(a), we show the training process where we train GNN models to predict parallelism opportunity and the parallel pattern.
Then, in Figure ~\ref{fig:workflow}(b), we show how at inference time, GNN is used to create context for the GNN-guided OMP prompt to guide the LLM to generate better parallel code.





\vspace{-6pt}
\subsection{Training}

The first step in our approach is training a Graph Neural Network to learn the features of the input programs.
\vspace{-4pt}
\subsubsection{Data Collection and Preprocessing}

First, we collect data to train our neural network to detect parallelism and patterns.
We want our neural network model to be able to realize if a region of a code (such as a loop) is parallelizable or not. We use the OMP\_Serial dataset ~\cite{chen2023learning} for this purpose. Also, some pre-processing is applied to transform the dataset into a graph representation of programs so that our GNN-based models can learn efficiently from the representation. Section \ref{sec:experimental_results} provides more details regarding the dataset and preprocessing. 

\vspace{-3pt}
\subsubsection{Program Representation}

While different program representations can be used to train neural networks, we use \perfograph ~\cite{tehranijamsaz2023perfograph} in this work as it incorporates control, call and data flow information of source programs. Also, \perfograph \xspace{} can represent multi-dimensional arrays and vectors in the programs. Additionally, it is numerically aware, meaning it can encode numbers. 
Experiments that have been conducted on \perfograph, show that this representation is effective for the task of parallelism discovery and pattern detection~\cite{tehranijamsaz2023perfograph}. We provide more details regarding structure of \perfograph \xspace{} and how node and edge embeddings are generated in Appendix ~\ref{sec:appendix-perfograph}.

\vspace{-4pt}
\subsubsection{Graph Neural Network (GNN) Training}

The benefit of using GNN is that programs can be represented as graphs, allowing to explicitly model various flows necessary for parallelism-related tasks. Using the \perfograph \xspace representation, we train a Graph Neural Network (GNN) to learn the flow-aware features of the programs specifically. We provide more details in Section \ref{sec:experimental_results}.

\vspace{-5pt}
\subsection{Inference}

In this part, we explain how GNNs are utilized to guide the LLMs to generate appropriate parallel code.

\subsubsection{Prompt Engineering}
In the first step, before even constructing the prompt for the LLM, we use GNN to identify if there is a parallelism opportunity in the given code. 
If there is no parallelism opportunity, the parallel version of the given code will not be generated. Figure ~\ref{fig:workflow}(b) shows the process of generating GNN-guided OMP prompt with relevant parallelization context, which is used to generate parallel OpenMP code. As said, the prompt would be used only if our GNN model predicts a parallelization opportunity.
Thereafter, the corresponding patterns will be predicted by the GNN as well. The supported patterns at the moment are: \texttt{do-all}, \texttt{private}, \texttt{reduction}, and \texttt{reduction and private} together. The \texttt{clause} placeholder in the prompt will be replaced by the name of the predicted pattern. We also designed a few-shot-COT OMP prompt for comparing with our approach manually by carefully selecting 5 samples that cover all parallelization cases described in Section \ref{sec:background}. 
We use this prompt for all our few-shot-COT experiments. The complete prompt is given in Appendix ~\ref{sec:appendix-few-shot}. Additionally, we conducted experiments using randomly selected samples for the prompt and the results are presented in Appendix \ref{appendix:randomized_few_shot_prompting}. Also, the Zero-shot-prompt is given in Appendix ~\ref{sec:appendix-zero-shot}. The choice of LLM depends on the user's preference. We experimented with two closed-source LLMs: GPT-4 and GPT-3.5, and two open-source LLMs: CodeLlama-34B ~\cite{roziere2023code} and CodeGen-16B ~\cite{nijkamp2022codegen}. The closed-source LLMs are accessed using the OpenAI GPT-3.5 and GPT-4 APIs ~\cite{OpenAI}.




\subsection{\ourscore}
\vspace{-10pt}
\begin{figure}[h]
    \centering
    \includegraphics[width=1\linewidth]{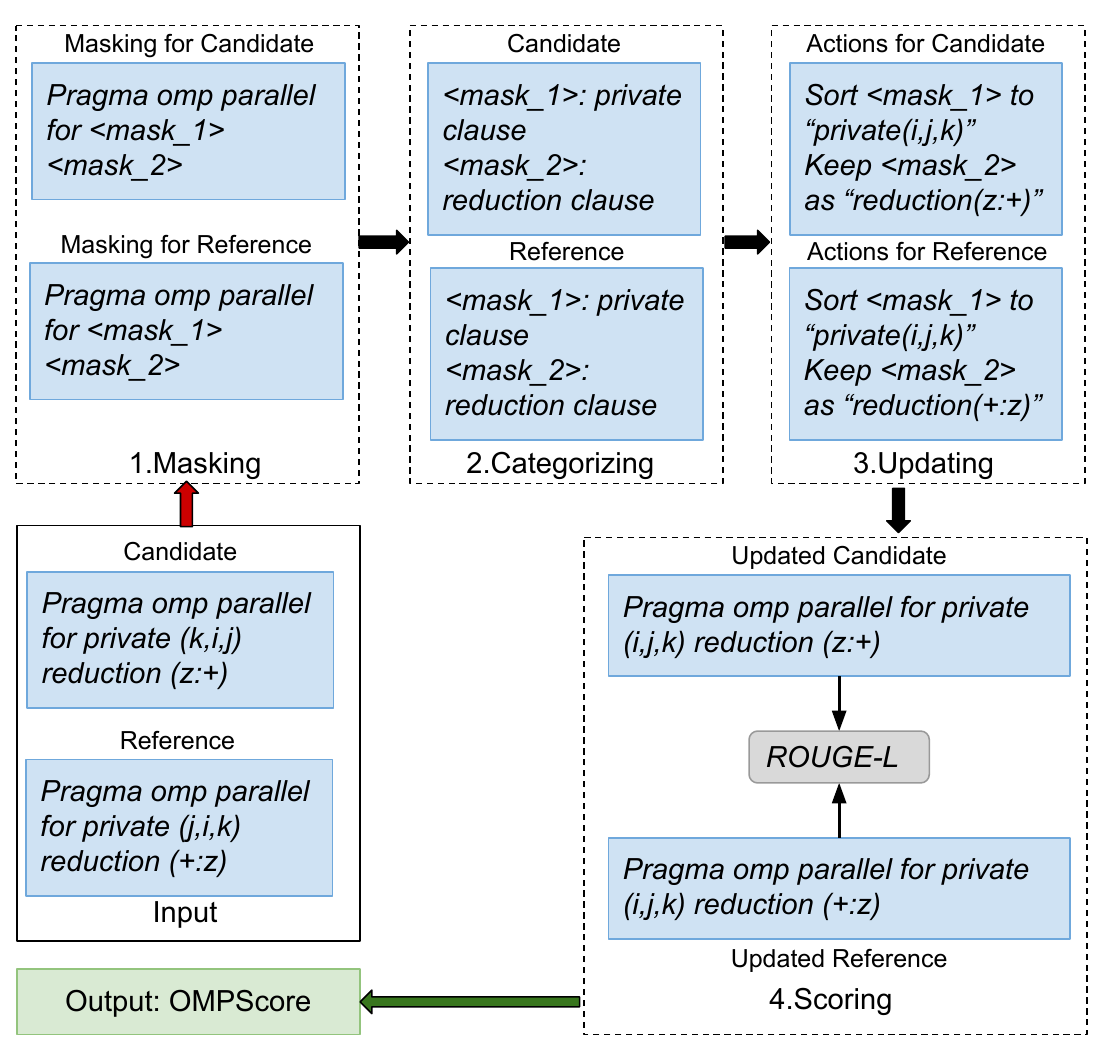}
    \caption{Overview of \ourscore.}
    \label{fig:ompScore}
\end{figure}

Some characteristics of OpenMP directives and clauses challenge the evaluation using existing textual similarity metrics. To illustrate, consider Figure ~\ref{fig:ompScore}, where we have a candidate and a reference directive, each composed of multiple clauses. One characteristic pertains to the order of variables or operands within certain clauses, where variable rearrangements may not alter semantic meaning (e.g., \texttt{private(k,j,i)} is equivalent to \texttt{private(j,i,k)}). Another characteristic involves specific clause types where the order of elements significantly affects directive performance. For example, in Figure ~\ref{fig:ompScore}, the candidate directive's \texttt{reduction(z:+)} clause is considered a mismatch to the reference directive. In essence, OpenMP directives encompass both order-sensitive and order-insensitive clauses, making a uniform treatment of all clauses as either order-sensitive or order-insensitive inadequate for accurate scoring. We introduce \ourscore, a metric that enhances Rouge-L score evaluation for \ourscore \xspace directives through a combination of regular expressions and program analysis. \ourscore \xspace comprises four key modules designed to preprocess input, both candidate and reference directives, to provide improved arguments for the Rouge-L score. In the initial stage, the Masking module detects potential clauses or directives for updating. It does so by identifying clauses within OpenMP directives using regular expressions initiated by OpenMP keywords (e.g., \texttt{private}, \texttt{shared}, or \texttt{reduction}) followed by open/close parentheses. Subsequently, the second module categorizes all identified masked spans based on the first word within each span, thereby determining the clause type (e.g., \texttt{private}, \texttt{shared}, or \texttt{reduction}). In the third module, we update the clauses considering two factors related to element order within OpenMP clauses. For instance, the \texttt{private} clause in the directive undergoes a sorting action while the \texttt{reduction} clause remains unchanged. To determine whether specific clause types are order-sensitive or order-insensitive, we refer to the official OpenMP documentation and articles\footnote{https://www.openmp.org/resources/tutorials-articles/}. For order-insensitive clause types, like the \texttt{private} clause, we alphabetically sort the elements within their respective element lists. Finally, in the fourth module, the updated candidate and reference directives serve as input for the Rouge-L scoring function, yielding the \ourscore, quantifying the similarity between the candidate and reference.

\vspace{-10pt}

%% file: sections/exp_results.tex
\section{Experimental Results}\label{sec:experimental_results}

We evaluate the effectiveness of \ourtool \xspace on several applications. In this section, we describe the details of those experiments. Also, we describe the components of \ourtool \xspace{} in detail. All deep learning models are run on computing nodes with the same configuration. Each computing node has an Intel Xeon Gold 6244 CPU with 32 cores and 366 GB of RAM.

\vspace{-4pt}
\subsection{Experimental Setup}

We use the OMP\_Serial dataset~\cite{chen2023learning} to pre-train the parallelism discovery and pattern detection module of \ourtool \xspace{}. 
\vspace{-3pt}
\subsubsection{Parallelism Detection Module}

The first step is to train our parallelism detection model to identify whether a region, such as a loop, can be executed in a parallel manner.
The OMP\_Serial dataset contains around 6k compilable C source files that are crawled from Github and well-known benchmarks like PolyBench ~\cite{pouchet2017polybench}, Starbench~\cite{andersch2013benchmark}, BOTS ~\cite{duran2009barcelona}, and the NAS Parallel Benchmark ~\cite{jin1999openmp}. However, since we use NAS Parallel Benchmark for evaluating the generated parallel codes, we carefully exclude all samples of NAS Parallel Benchmark from the dataset for our pre-training phase so that our model does not "see" those samples beforehand. After exclusion, LLVM Intermediate Representations (LLVM IR) of source C files are generated. To augment the dataset and increase the size of training data, we compile programs using different LLVM optimization flags following the approach of ~\cite{tehranijamsaz2022learning}. Ultimately, we have around 10k IR files (6041 \texttt{parallel}, 4194 \texttt{non-parallel}).

\vspace{-3pt}
\subsubsection{Pattern Detection Module}

The OMP\_Serial dataset also contains 200 \texttt{private} and 200 \texttt{reduction} loops. However, after removing samples taken directly from NAS benchmark and extracted templates, around 158 \texttt{private} and 137 \texttt{reduction} samples are left. Finally, we apply LLVM optimization flags similarly as mentioned above and generate around 4k IR files (2160 \texttt{private}, 2100 \texttt{reduction}). The \texttt{private} clause detection model determines the need for a \texttt{private} clause. Similarly, the \texttt{reduction} clause detection model is used to identify whether we need a \texttt{reduction} clause in the OpenMP directive or not. For training \texttt{private} clause detection model, two classes are created: \texttt{private} (2160 files) and \texttt{non-private} (contains 2000 files, 50\% of those are taken randomly from \texttt{reduction} and 50\% of those are randomly taken from \texttt{non-parallel}). Similarly, for training \texttt{reduction} clause detection model, two classes are created: \texttt{reduction} (2100 files) and \texttt{non-reduction} (contains 2000 files, 50\% of those are taken randomly from \texttt{private} and 50\% of those are randomly taken from \texttt{non-parallel}). These two models make up the parallel pattern detection module of \ourtool.

\vspace{-4pt}
\subsubsection{GNN Classifier}

We use the DGL-based~\cite{wang2019deep} implementation of RGCN ~\cite{schlichtkrull2018modeling} with 6 GraphConv layers for all three GNN models. Each source program is a heterogeneous graph represented by \perfograph ~\cite{tehranijamsaz2023perfograph}, so the HeteroGraphConv module in each layer is used with the `sum' aggregation function. All 3 models are trained for 120 epochs, and the checkpoints with the highest validation accuracy is saved for later inference. The model hyperparameter details, loss curves, and training times are reported in Appendix ~\ref{sec:appendix-loss--training-times}. Table \ref{nas-rodinia-parallel-pattern} shows the accuracy of the 3 GNN models used for context generation.

\begin{table}[]
    \caption{Accuracy of Parallelism Detection, Private Detection and Reduction Detection Models.}
    \vspace{-12pt}
    \label{nas-rodinia-parallel-pattern}
    \begin{center}
    \setlength\tabcolsep{1pt}
    \resizebox{1\columnwidth}{!}{%
    \begin{tabular}{ccc}
    \multicolumn{1}{c}{\bf Model} &\multicolumn{1}{c}{\bf NAS Benchmark} &\multicolumn{1}{c}{\bf Rodinia Benchmark} \\ \hline
    Parallelism Discovery Accuracy &          94.44\%   &       100\% \\ 
    Private Detection Accuracy      &          92.86\%   &      100\% \\ 
    Reduction Detection Accuracy   &          100\%     &       100\%
    \vspace{-34pt}
    \end{tabular}%
    }
    \end{center}
\end{table}

\subsubsection{Inference and GNN-based Prompt Generation}

For inference, the three pre-trained models are applied sequentially. First, the input code is passed to the parallelism detection model. If it classifies a loop as \texttt{parallel}, then it is passed to the \texttt{private} clause detection model. If the second model classifies it as a \texttt{private} loop, then the \texttt{private} clause is added to the OMP prompt. Finally, the loop is passed to the \texttt{reduction} clause detection model, and similarly, if it classifies the loop as a \texttt{reduction} loop, the \texttt{reduction} clause is also added to the OMP prompt (Figure ~\ref{fig:workflow}). 

\vspace{-3pt}
\subsubsection{Generating OpenMP Clauses and Parallel Codes}

After creating the GNN-guided OMP prompts, the LLMs are invoked to generate the parallel counterpart of the sequential programs. We use four LLMs to demonstrate the performance of \ourtool; note that for the LLMs, the \textit{temperature} parameter is set to zero to make the models deterministic in predicting the OpenMP constructs. We evaluate the performance of \ourtool \xspace{} on 11 applications of two benchmarks: NAS Parallel Benchmark and Rodinia Benchmark ~\cite{che2009rodinia}. These applications are developed targeting HPC platforms and heterogeneous computing. Both of the benchmarks have OpenMP annotated loops and their sequential version from experienced developers. 

\vspace{-3pt}
\subsubsection{Evaluation}

To evaluate the quality of the generated codes, we use CodeBERTScore ~\cite{zhou2023codebertscore} and also metrics (ParaBLEU ~\cite{wen2022babeltower}, \ourscore) that are specifically designed for evaluating parallel codes. However, ParaBLEU is specifically designed to evaluate CUDA code. Hence, we modified ParaBLEU score following the same idea of ~\cite{wen2022babeltower}. We provide the details of the implementation in Appendix ~\ref{sec:appendix-ParaBLEU}.

\begin{table}[h]
\vspace{-5pt}
\caption{Application-wise extracted loops count for NAS and Rodinia benchmark in the testing set}
\vspace{-10pt}
\begin{center}
\resizebox{0.7\columnwidth}{!} {%
\begin{tabular}{ccc}
\hline
\textbf{Benchmark}       & \textbf{Application} & \textbf{Number of loops} \\ \hline
\multirow{9}{*}{NAS}     & BT                   & 7                        \\
                         & IS                   & 6                        \\
                         & CG                   & 10                       \\
                         & FT                   & 5                        \\
                         & EP                   & 6                        \\
                         & LU                   & 13                       \\
                         & MG                   & 15                       \\
                         & SP                   & 28                       \\ \cline{2-3} 
                         & \textbf{Total}       & \textbf{90}              \\ \hline
\multirow{5}{*}{Rodinia} & BFS                  & 1                        \\
                         & B+ Tree              & 6                        \\
                         & Heartwall            & 13                       \\
                         & 3D                   & 1                        \\ \cline{2-3} 
                         & \textbf{Total}       & \textbf{21}              \\ \hline
\end{tabular}
}
\end{center}
\label{tbl:nas-rod-loops}
\vspace{-8pt}
\end{table}

\vspace{-7pt}


\subsection{Evaluating Code Generation on NAS Parallel Benchmark}
\begin{figure}[h]
    \centering
    \includegraphics[width=1\linewidth]{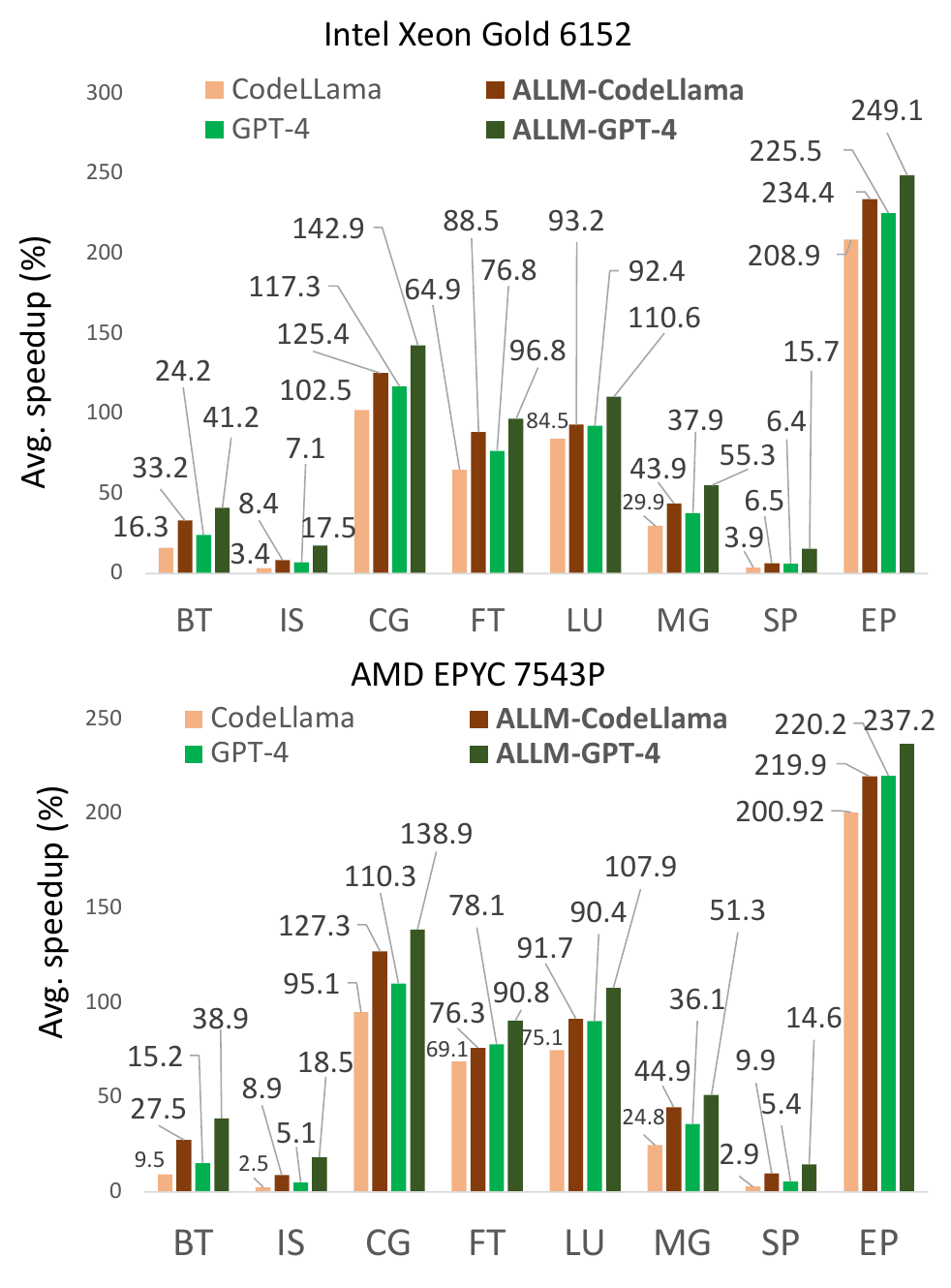}
    \vspace{-25pt}
    \caption{Speedup gain across individual applications in NAS Parallel Benchmark. ALLM-GPT-4 achieves max \textbf{24.7\%} and  \textbf{28.6\%} better speedup than GPT-4 for CG in Intel and AMD cpus, respectively.}
    \vspace{-20pt}
    \label{fig:nas-ind-speedup}
\end{figure}

\begin{table}[]
    \caption{Results on NAS Parallel Benchmark Suite. Higher indicates better. 100 score means a perfect match with the ground-truth values. Bold fonts indicate better scores.}
    \vspace{-10pt}
    \label{nas-scores}
    \begin{center}
    \setlength\tabcolsep{2pt}
    \resizebox{1\columnwidth}{!} {%
    \begin{tabular}{ccccccccc}
    \multicolumn{1}{c}{\bf Model} &\multicolumn{1}{c}{\bf CBTScore} &\multicolumn{1}{c}{\bf ParaBLEU} &\multicolumn{1}{c}{\bf OMPScore} \\ \hline
    0-shot-COT-CodeGen-16B      & 72.8 & 17.43  & 48.97 \\
    few-shot-COT-CodeGen-16B    & 73.15 & 21.77  & 56.19 \\
    \bf \ourtool-CodeGen-16B    & \bf 83.8 & \bf 26.29 & \bf 65.3 \\ \hline
    0-shot-COT-CodeLlama-34B    & 74.0 & 26.48   & 45.44\\
    few-shot-COT-CodeLlama-34B  & 78.6 & 37.46  & 60.52\\
    \bf \ourtool-CodeLlama-34B  & \bf 96.0 & \bf 47.73  & \bf 94.46 \\ \hline
    0-shot-COT-GPT-3.5          & 72.3 & 27.44 & 41\\
    few-shot-COT-GPT-3.5        & 74.2 & 30.72 & 50.71\\
    \bf \ourtool-GPT-3.5        & \bf 95.2 & \bf 48.28 & \bf 95.15 \\ \hline
    0-shot-COT-GPT-4            & 73.8 & 27.49  & 46.4\\
    few-shot-COT-GPT-4          & 76.5 & 29.97  & 58.06\\
    \bf \ourtool-GPT-4          & \bf 96.4 &  \bf 48.48  & \bf 95.15 \\
    
    \end{tabular}
    }
    \end{center}
\end{table}

First, we evaluate \ourtool \xspace{} on NAS Parallel Benchmark by extracting loops containing OpenMP pragmas from the eight applications. For loop extraction, we first annotate the loops using the Rose outlining tool ~\cite{quinlan2011rose}. Then, we compile and generate the IR for the outlined code. Finally, the \texttt{llvm-extract} command is used to extract the loop-specific IR from the full IR. A total of 454 loops (\texttt{private}: 264, \texttt{reduction}: 17, \texttt{non-parallel}: 173) are extracted. 80\% of the loops are used to fine-tune our pretrained GNN models. The GNN models are trained for 120 epochs. Then 20\% of loops (90 loops) are used for evaluating \ourtool. Of those 90 loops, 58 are \texttt{parallel}, with 56 loops having \texttt{private} clause and two loops having \texttt{reduction} clause. The rest 32 loops are \texttt{non-parallel}. \ourtool \xspace{} achieves 94.44\% accuracy in parallelism discovery by correctly predicting 55 out of 58 \texttt{parallel} loops and 30 out of 32 \texttt{non-parallel} loops. Also, \ourtool \xspace{} correctly detects 52 out of 56 loops with \texttt{private} clause, and it correctly detected all two loops with \texttt{reduction} clause. Table ~\ref{tbl:nas-rod-loops} shows the loops that are extracted from different applications of NAS benchmark. In Table ~\ref{nas-scores}, we compare the performance of the codes generated by using the basic OMP prompt and GNN-guided OMP prompt (denoted as \ourtool-LLM-name in all tables). We use different score metrics as well as OMPScore for the comparison, and it can be observed that our \ourtool \xspace{} approach improves all LLMs in terms of these metrics scores. For example, \ourtool \xspace{} augmented GPT-4 (\ourtool-GPT-4) can improve the baseline GPT-4 by 19.9\% in terms of CodeBERTScore and 37.09\% in terms of \ourscore, which is more appropriate for the evaluation of generated OpenMP configurations (Table ~\ref{nas-scores}). 

Apart from the metric scores, we also evaluate the performance of the generated codes by measuring their execution time. We replace the sequential loops in the testing set with the parallel loops generated using both regular LLMs and \ourtool \xspace{} augmented LLMs and then execute the application five times to measure average execution time. The speedup over sequential version is then calculated using Equation \ref{eq:speedup-eq-1} for both regular and \ourtool \xspace{} augmented LLM generated parallel version.
\vspace{-10pt}
\begin{equation}
    Speedup\% = (\frac{Avg. Sequential Runtime}{Avg. Parallel Runtime}-1)*100
    \label{eq:speedup-eq-1}
\end{equation}

To test the robustness of \ourtool \xspace{} across different hardware the runtime experiments are performed on two different CPU architectures (Intel \& AMD). It is observed from Figure \ref{fig:nas-ind-speedup} that for all 8 applications in NAS Parallel Benchmark, \ourtool \xspace guidance resulted in significant improvement in speedup than base LLMs (GPT-4 \& CodeLlama-34B). The results in Figure \ref{fig:nas-ind-speedup} are reported for 4 threads. We report the detailed results of the runtime experiments like Input (\ref{sec:appendix-input-nas},
\ref{sec:appendix-input-rodinia}), CPU architecture details (\ref{ref:hardware-spec}) and scalability testing with different number of threads (\ref{sec:appendix-thread-config}) in Appendix. Due to the high computation cost of executing the applications we only considered the codes generated by few-shot setting for the base LLMs as it generated better codes based on the findings from Table ~\ref{nas-scores}.

\vspace{-4pt}
\subsection{Evaluating Code Generation on Rodinia Benchmark}
\begin{figure}[t]
    \centering
    \includegraphics[width=1\linewidth]{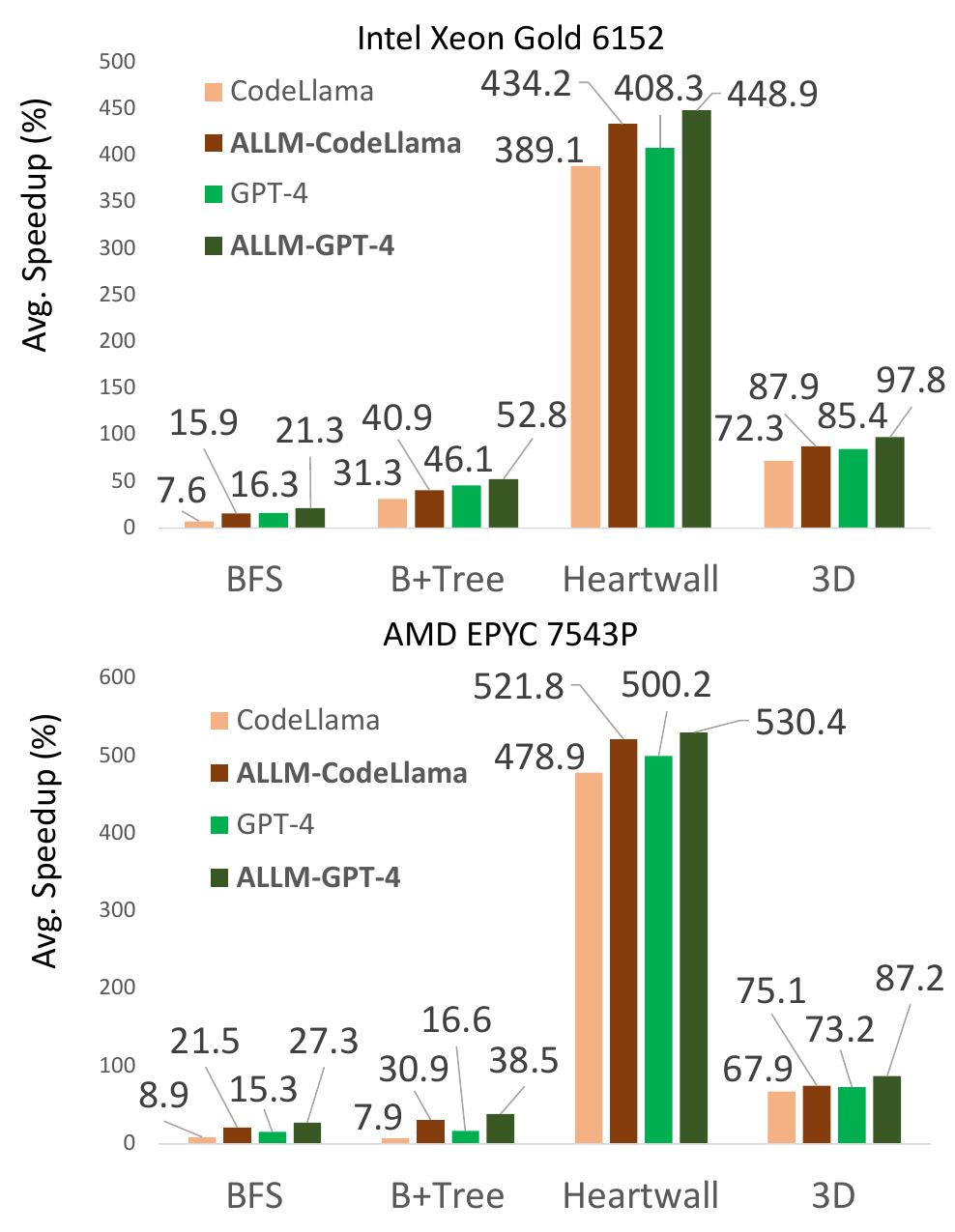}
    \vspace{-25pt}
    \caption{Speedup gain across individual applications in Rodinia-3.1 Benchmark. ALLM-GPT-4 achieves max \textbf{40.6\%} and  \textbf{30.2\%} better speedup than GPT-4 for Heartwall in Intel and AMD cpus, respectively.}
    \vspace{-14pt}
    \label{fig:rod-ind-speedup}
\end{figure}
We further apply \ourtool \xspace{} on four applications of Rodinia Benchmark that our GNN models have not seen at all. These applications are developed targeting heterogeneous computing. We extracted 21 loops from these applications using the method described earlier. Out of these 21 loops, 15 loops contain the \texttt{private} clause, and 6 loops contain the \texttt{reduction} clause. \ourtool \xspace{} is applied to detect parallelism and pattern of these 21 loops. Out of the 21 loops, \ourtool \xspace{} is able to correctly detect and classify all the 15 loops with \texttt{private} clauses and 6 loops with \texttt{reduction} clauses. Table ~\ref{tbl:nas-rod-loops} shows the loops that are extracted from different applications of Rodinia benchmark. Table ~\ref{rodinia-scores} shows the results. We can see that \ourtool \xspace{} guidance results in better code generation in terms of all the considered metrics. For example,  \ourtool \xspace{} augmented GPT-4 (\ourtool-GPT-4) can improve the baseline GPT-4 by 6.48\% in terms of CodeBERTScore and 9.09\% in terms of \ourscore. It can be seen that \ourtool \xspace{} has a significantly higher \ourscore \xspace{} for this dataset, too. Runtime experiments are done in a similar manner for Rodinia-3.1 on Intel and AMD cpus. It is observed from Figure \ref{fig:rod-ind-speedup} that for all 4 applications in Rodinia-3.1 Benchmark, \ourtool \xspace guidance resulted in significant improvement in speedup than base LLMs (GPT-4 \& CodeLlama-34B). The results in Figure \ref{fig:rod-ind-speedup} are also reported for 4 threads. We provide more details of runtime experiments in Appendix \ref{ref:hardware-spec}, \ref{sec:appendix-input-rodinia}, and \ref{sec:appendix-thread-config}.


\begin{table}[]
    \caption{Results on Rodinia-3.1 Benchmark Suite. 100 score means a perfect match with the ground-truth values. Bold fonts indicate better scores.}
    \vspace{-12pt}
    \label{rodinia-scores}
    \begin{center}
    \setlength\tabcolsep{2pt}
    \resizebox{1\columnwidth}{!} {%
    \begin{tabular}{ccccccccc}
    \multicolumn{1}{c}{\bf Model}  &\multicolumn{1}{c}{\bf CBTScore} &\multicolumn{1}{c}{\bf ParaBLEU} &\multicolumn{1}{c}{\bf OMPScore} \\ \hline

    0-shot-COT-CodeGen-16B     & 79.0 & 28.52  &52.88 \\
    few-shot-COT-CodeGen-16B   & 79.8 & 31.44  &55.91 \\
    \bf \ourtool-CodeGen-16B   & \bf 82.2 & \bf 41.73  & \bf 64.57 \\ \hline
    0-shot-COT-CodeLlama-34B   & 85.9 &  41.92  &  69.55\\
    few-shot-COT-CodeLlama-34B & 88.2 &  46.13  &  76.92\\
    \bf \ourtool-CodeLlama-34B & \bf 96.4 & \bf 66.13  & \bf 95.27 \\ \hline
    0-shot-COT-GPT-3.5         & 91.0 &  46.98  & 79.37 \\
    few-shot-COT-GPT-3.5       & 91.8 &  51.08  & 88.13 \\
    \bf \ourtool-GPT-3.5       & \bf 97.6 &  \bf 68.36  &  \bf 98.1 \\ \hline
    0-shot-COT-GPT-4           & 91.2 & 49.14 & 79.37\\
    few-shot-COT-GPT-4         & 92.12 & 54.43 & 89.01\\
    \bf \ourtool-GPT-4         & \bf 98.6 & \bf 69.09  & \bf 98.1 \\
     \vspace{-8pt}
    \end{tabular}
    }
    \end{center}
    \vspace{-27pt}
\end{table}

\vspace{-5pt}

\subsection{Increasing Developer Productivity}

\ourtool \xspace can greatly increase developer efficiency in writing parallel codes by eliminating cases where parallelization is not possible. For example, in the NAS benchmark \ourtool \xspace correctly detected 30 out of 32 \texttt{non-parallel} loops. As there are 90 loops in the benchmark, it means a developer can safely filter out 33.33\% of the loops from parallelization consideration. We provide more details regarding how to handle FP and FN efficiently in Appendix ~\ref{sec:appendix-correctness}.

\subsection{Human Evaluation Results}

For human evaluation, we manually evaluate the 90 loops in NAS benchmark that are generated using the \ourtool-GPT-4. For each loop, we compare the generated OpenMP directives with the original ones, which are considered the ground truth values. Each loop is examined by two independent observers and they allocate scores ranging from 0 (low-quality) to 5 (high-quality) to the predicted directive based on the number of operations needed to transform the predicted directive into the original directive. Each modification operation results in a deduction of 1 point from the score. The score for a particular loop parallelized using \ourtool-GPT-4 is the average score of the two observers. Then we calculate the average score of the 90 loops, and on average \ourtool-GPT-4 achieved an impressive 4.72/5 on the 90 evaluated loops on NAS, which indicates a 94.4\% match with the ground truth values. On the same 90 loops, \ourtool-GPT-4 achieves \ourscore \xspace of 95.15 (Table \ref{nas-scores}) which is very close to the human evaluation score indicating the effectiveness of \ourscore \xspace to correctly evaluate the generated parallel code.

\subsection{Comparing with Traditional and Deep Learning Approaches}

We further compare the parallelism discovery of \ourtool \xspace{} against SOTA traditional and Deep Learning-based approaches in Appendix ~\ref{sec:appendix-traditional} and Appendix  \ref{appendix:compare-deep-learning}. Results show that \ourtool \xspace significantly outperforms both SOTA traditional and Deep Learning-based approaches.

\subsection{Extending beyond OpenMP}

We also evaluate \ourtool \xspace on another widely used parallel programming model OpenACC. OpenACC supports a lot of the parallel configurations offered by OpenMP, and it also has some benchmarks that we can use as ground-truths to evaluate \ourtool. For this experiment, two OpenACC benchmarks: EPCC Benchmark \cite{johnson2013epcc} and PolyBench-OpenACC Benchmark \cite{grauer2012auto}, are used. A total of 34 loops are extracted from these two benchmarks. These include 22 \texttt{parallel} (10 \texttt{private}, 12 \texttt{reduction}) and 12 \texttt{non-parallel} loops. We utilize the pre-trained GNN modules of \ourtool \xspace and fine-tune for up to 120 epochs using 50\% of the total loops to ensure that \ourtool \xspace adapts to the new parallel programming framework OpenACC. The remaining 50\% is used for testing \ourtool's ability to predict parallelization patterns in OpenACC. The test set contains 12 \texttt{parallel} (5 \texttt{private}, 7 \texttt{reduction}) and 5 \texttt{non-parallel} loops. Table \ref{openacc-results} shows that \ourtool \xspace can correctly detect all \texttt{parallel} and \texttt{non-parallel} loops as it has an accuracy of 100\% for the parallelism discovery task. For each \texttt{private} and \texttt{reduction} detection task, there is only one mismatch, and they have an accuracy of 80\% and 85.71\%, respectively. We discuss these mismatches in detail in Appendix \ref{sec:appendix-openacc-mismatch}. However, it can be observed that \ourtool \xspace has a good overall accuracy of 88.24\%.

\begin{table}[!ht]
    \caption{\ourtool \xspace results on OpenACC parallelization patterns}
    \label{openacc-results}
    \begin{center}
    \vspace{-14pt}
    \setlength\tabcolsep{4pt}
    \resizebox{1\columnwidth}{!}{%
    \begin{tabular}{ccc}
    \multicolumn{1}{c}{\bf Tasks} &\multicolumn{1}{c}{\bf Accuracy} &\multicolumn{1}{c}{\bf Correct / \# Samples} \\ \hline
    Parallelism Discovery  &          100.00\%   &       17 / 17 \\ 
    Private Detection      &          80.00\%   &      4 / 5 \\ 
    Reduction Detection   &          85.71\%     &       6 / 7  \\ \hline
    \textbf{Overall Accuracy}     &     \textbf{88.24\%}
    \vspace{-28pt}
    \end{tabular}%
    }
    \end{center}
\end{table}

\begin{figure*}[!ht]
    \centering
    \includegraphics[width=1\linewidth]{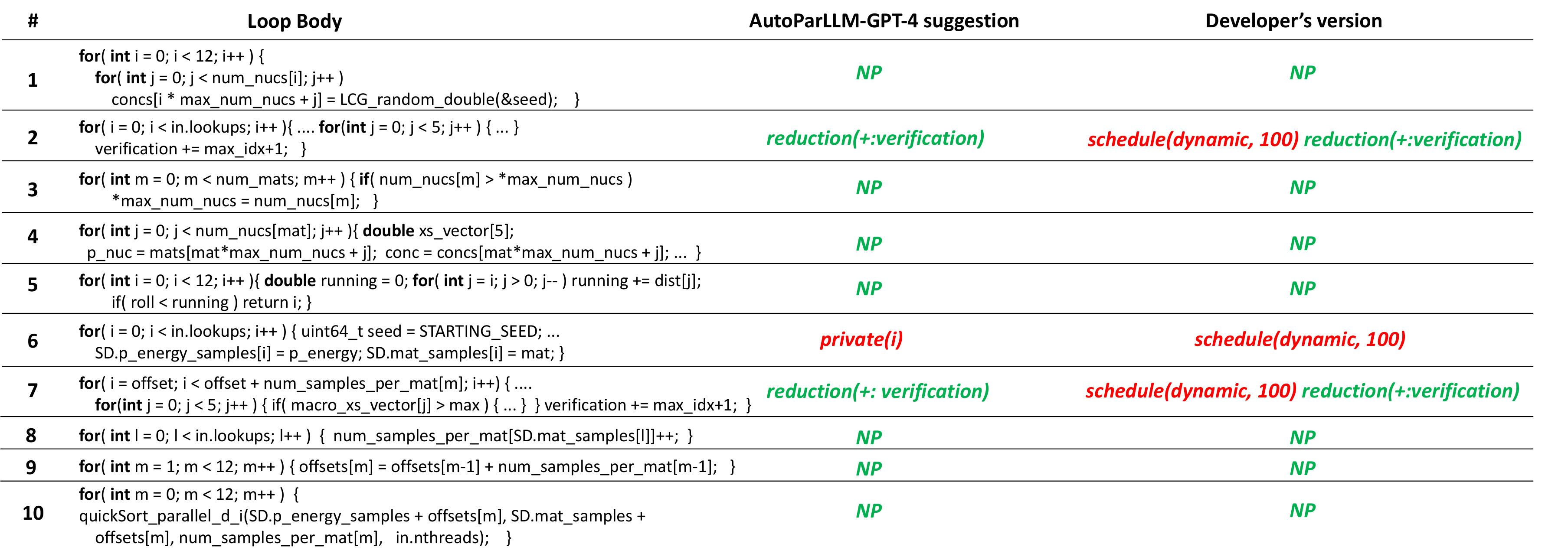}
    \vspace{-12pt}
    \caption{Use Case Analysis Results for XSBench. Outermost loops are considered. \textbf{NP = Non-Parallel}. \textbf{Green cases} show a match, whereas \textbf{Red cases} show a mismatch. For parallel loops, the phrase \textbf{`\#pragma omp parallel for'} is omitted for simplification. Loops are simplified.}
    \vspace{-17pt}
    \label{fig:xsbench}

\end{figure*}

\subsection{Use Case Analysis} \label{section-appendix-use-case}


For analyzing the effectiveness of \ourtool \xspace on a real-world HPC application, XSBench is chosen as it is considered a useful tool for performance analysis of High-Performance Computing systems ~\cite{Tramm:wy} and also OpenMP compatible. For this experiment, we used our \ourtool-GPT-4 configuration. Firstly, the loops from the XSBench are extracted. Then, the \perfograph \xspace{} representation of the loops are constructed and passed to the GNN-based predictors of \ourtool. The predictors provide useful feedback regarding the loops, which is ultimately used by \ourtool-GPT-4 to parallelize the loops. Figure ~\ref{fig:xsbench} shows the detailed results. All non-parallel codes are correctly detected. Also \ourtool-GPT-4 \xspace{} is able to correctly generate all the \texttt{reduction} clauses along with the \texttt{reduction} operator and variable. \ourtool-GPT-4 \xspace{} failed to generate the \texttt{schedule} clauses which is expected as \ourtool \xspace{} is not trained to generate the \texttt{schedule} clause. Also, the \texttt{schedule} clause is configurable, meaning developers can try different chunk sizes and scheduling techniques (\texttt{static}, \texttt{dynamic}) to obtain optimal performance. Finally, it can be observed that there is a mismatch in the \texttt{private} clause. However, the variable \texttt{i} is the loop counter, and \ourtool-GPT-4 \xspace{} is actually right in predicting the \texttt{private} clause and also the \texttt{private} variable \texttt{i}. However, as the loop counters are considered \texttt{private} by default in OpenMP, the developers sometimes omit to decorate these types of loop counter variables explicitly with a \texttt{private} clause, although it is considered good practice. 
\vspace{-2.03pt}

%% file: sections/related_works.tex
\section{Related Works}
\vspace{-4pt}
We describe the related works in detail in Appendix ~\ref{sec:appendix-related-works}. Here, we discuss works that are more closely related to our study. There are DL-based works that focuses on parallelization \cite{chen2023learning, harellearning, shen2021towards, shen2023multigraph, schaarschmidt2021automap}. These works mostly predict parallelism opportunities, but they do not generate complete parallel code. Finetuning of LLMs has also been explored for OpenMP-based parallelization  \citep{chen2024ompgpt}. However, the results indicate that the accuracy of the generated parallelization clauses is quite low. In contrast to these prior works, our approach uniquely combines the strengths of GNNs with LLMs to generate prompts enriched with relevant context. This hybrid method leverages the advantages of both GNNs and LLMs, providing a versatile solution compatible with any LLM. Also as discussed earlier a large body of works use ICL to enhance the performance of LLMs. ICL has been utilized in both the training \cite{min2021metaicl, wei2023symbol, gu2023pre, iyer2022opt} and inference stages \cite{li2023finding, wang2024large, wang2022self, li2023mot, xu2023small, fu2022complexity, wei2022chain, zhang2022automatic}. In these studies, context involves providing the model with sample inputs and corresponding expected responses before the test inputs. Whereas, we proposed a novel approach for generating context of input codes using GNN-based guidance.

\vspace{-5pt}


%% file: sections/conclusion.tex
\vspace{-5pt}
\section{Conclusion}
\vspace{-7pt}

\ourtool \xspace{}, sits on top of GNNs and LLMs to enable automatic parallelization of programs. We developed \ourtool \xspace{} as an intelligent parallelism assistant for developers but not as a replacement. Based on our results, we believe it is fair to say that \ourtool \xspace{} generates parallelized versions of loops with very little human effort and greatly increases developers’ efficiency. 

\vspace{-10pt}